\documentclass{article}

\usepackage[final]{corl_2026} 

\usepackage{amsmath}   
\usepackage{makecell} 
\usepackage{amsfonts}  
\usepackage{mathtools} 
\usepackage{booktabs}

\usepackage{wrapfig}
\usepackage{float}

\newcommand{\dplus}{D$+$}
\newcommand{\dminus}{D$-$}

\usepackage{graphicx}  
\usepackage{hyperref}  
\usepackage{multirow}
\usepackage[table]{xcolor}
\usepackage{tabularx}
\usepackage{bbm}

\usepackage{amsmath,amsfonts,bm}

\def\eqref#1{equation~\ref{#1}}

\def\1{\bm{1}}

\DeclareMathAlphabet{\mathsfit}{\encodingdefault}{\sfdefault}{m}{sl}
\SetMathAlphabet{\mathsfit}{bold}{\encodingdefault}{\sfdefault}{bx}{n}

\hypersetup{
  pdftitle={CoRE-VLA: Towards Scalable and Robust Vision-Language-Action Modeling via Conditional Routing of Experts},
  pdfauthor={Haozhe Zhang, Sixian Li, Yifei Zhang, Zezheng Huai, Hao Chen, Chunhua Shen, Jingjing Gong, Xipeng Qiu},
  pdfsubject={arXiv preprint},
  pdfkeywords={Vision-Language-Action Models, Robotic Manipulation, Conditional Routing, Sparse Computation, Multimodal Policy Learning}
}

\title{CoRE-VLA: Towards Scalable and Robust Vision-Language-Action Modeling\\ via Conditional Routing of Experts}

\author{
\parbox{0.95\textwidth}{
\centering
Haozhe Zhang$^{1,2}$,
Sixian Li$^{3,2}$,
Yifei Zhang$^{4,2}$,
Zezheng Huai$^{5,2}$,\\
Hao Chen$^{1}$,
Chunhua Shen$^{1}$,
Jingjing Gong$^{2, \dagger}$,
Xipeng Qiu$^{3,2, \dagger}$\\[0.5em]
{\normalfont\mdseries
$^{1}$Zhejiang University \quad
$^{2}$Shanghai Innovation Institute \quad
$^{3}$Fudan University \\
$^{4}$Nanjing University \quad
$^{5}$Jilin University\\[0.5em]
$^{\dagger}$Corresponding authors
}
}
}

\begin{document}
\maketitle

\begin{abstract}
Vision-language-action (VLA) models have advanced generalist robotic manipulation, yet real-world deployment reveals a fundamental challenge: robots are equipped with diverse and heterogeneous sensor configurations, auxiliary sensors can fail unexpectedly during operation, and different robot embodiments often lack certain sensors by design. A unified policy that can exploit auxiliary perceptual inputs when available while remaining reliable under sensor absence — whether incidental or by design — is therefore essential for practical deployment. However, existing VLA policies couple action generation to a fixed sensor set through shared dense computation, making them brittle when sensors are missing and limiting their ability to specialize across diverse tasks and long-horizon behaviors. We propose CoRE-VLA, a scalable and robust VLA framework that formulates action generation as context-conditioned sparse computation. Sensor availability gates modality-specialized experts, enabling graceful degradation under missing sensors without retraining. Task intent further routes action-side representations to task-relevant experts, improving specialization across diverse tasks and long-horizon subgoals. While the framework is designed to accommodate different auxiliary sensors, we focus on depth as a representative and practically important auxiliary modality in our experiments. Experiments on LIBERO, RoboCasa GR1 Tabletop, and real-world dual-arm manipulation show that CoRE-VLA achieves strong results on long-horizon and multi-task benchmarks, and outperforms both a dense-action-generator ablation and a strong pretrained VLA baseline, including in zero-shot generalization to unseen scenarios. Modality analysis shows that CoRE-VLA can exploit auxiliary depth when available while remaining robust when depth is unavailable during deployment.

\end{abstract}


\section{Introduction}

Vision-language-action (VLA) models have emerged as a promising paradigm for generalist robotic manipulation by coupling pretrained vision-language representations with continuous action generation~\cite{kim2025openvla,black2024pi_0,intelligence2025pi_,bjorck2025gr00t}.
Despite this progress, deploying VLA policies across real-world robotic platforms remains challenging because robots often have different sensor configurations.
Some platforms provide auxiliary perceptual signals such as depth, tactile sensing, or force feedback, while others lack these sensors due to differences in hardware design, configuration, or deployment constraints.
Even when such sensors are available, they may fail or become unreliable during operation.
A practical VLA policy should therefore exploit auxiliary perceptual inputs when they are available, while still acting reliably from RGB, proprioception, and language alone when they are absent.

Most existing VLA policies are trained and deployed with a fixed observation interface and a shared dense action generator.
This design tightly couples action generation to a fixed sensor set.
As a result, training a policy directly with auxiliary sensors can make it over-rely on them and degrade when those sensors are unavailable at deployment time.
This sensor coupling is especially problematic for real-world robotic systems, where sensor availability can differ across embodiments and may change during operation.

A second limitation is that a shared dense action generator applies the same computation path to heterogeneous tasks and long-horizon subgoals.
Manipulation tasks often require different forms of action processing, such as spatially precise reaching, object grounding, bimanual coordination, contact-rich interaction, and deformable-object handling.
Prior studies in multi-task learning show that forcing diverse objectives to share the same parameters can lead to conflicting gradients, task interference, and negative transfer~\cite{yu2020gradient,standley2020tasks}.
This challenge is especially relevant to long-horizon manipulation, where a single trajectory may involve distinct subgoals.
Thus, scalable VLA modeling requires not only stronger perception backbones, but also an action generator that can adapt its computation to both the current task intent and the available sensory context.
Moreover, because visual tokens typically dominate the multimodal token sequence in VLAs, 
directly deriving routing decisions from the pooled multimodal representation can dilute 
the instruction-level task intent that is most relevant for action specialization.

\begin{figure}[t!]
\centering
\includegraphics[width=\linewidth]{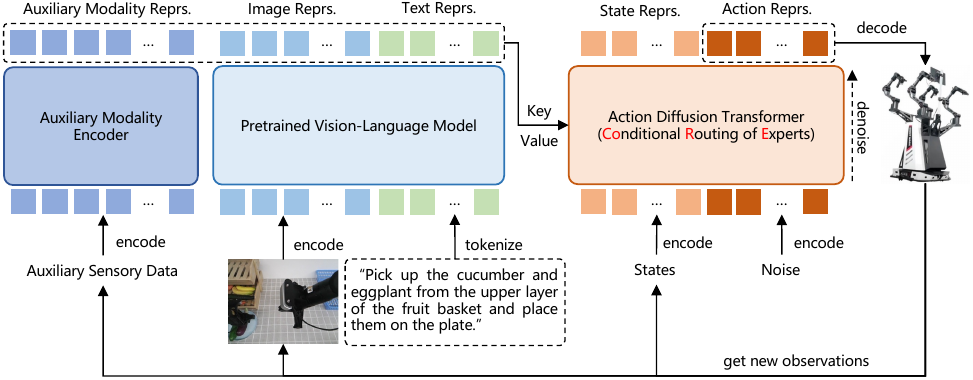}
\vspace{-15pt}
\caption{Overview of the CoRE-VLA architecture. CoRE-VLA encodes RGB observations and language instructions with a pretrained VLM, encodes auxiliary sensory inputs with a modality encoder, and conditions a flow-matching Action Diffusion Transformer on proprioceptive states and noisy action representations. See Figure~\ref{fig:details} for details of the CoRE block in the Action DiT. }
\vspace{-5pt}
\label{fig:framework}
\end{figure}

We propose \textbf{CoRE-VLA}, a scalable and robust VLA framework based on \textbf{Conditional Routing of Experts}, as shown in Figure~\ref{fig:framework}.
Our key idea is to formulate action generation as \emph{context-conditioned sparse computation}, rather than a fixed dense computation path shared by all tasks and sensor configurations.
CoRE-VLA uses two forms of context.
First, sensor availability gates modality-specialized experts, allowing the policy to exploit auxiliary perceptual cues when they are present while masking the corresponding experts when they are absent.
Second, to preserve an explicit task-level routing signal under visual-token-dominated 
multimodal representations, CoRE pools VLM-encoded text tokens as the task-intent 
representation and uses it to route action-side representations to task-relevant experts.
Together with modality dropout, this design allows an RGB-pretrained VLA policy to be extended with auxiliary perceptual modalities through continued training, while preserving reliable action generation when auxiliary sensors are unavailable.

While CoRE-VLA is designed to accommodate different auxiliary sensors, we focus on depth as a representative and practically important auxiliary modality in our experiments.
We evaluate CoRE-VLA on LIBERO, RoboCasa GR1 Tabletop, and real-world dual-arm manipulation tasks.
Across simulation benchmarks, CoRE-VLA achieves strong performance on multi-task and long-horizon manipulation.
In real-world deployment, CoRE-VLA outperforms both a dense action-generator ablation and a strong pretrained VLA baseline, including in zero-shot generalization to unseen fabric-folding scenarios.
Modality analysis further shows that CoRE-VLA can exploit auxiliary depth when available while maintaining reliable execution when depth is removed or unavailable.

Our contributions are threefold:
\begin{itemize}
\item To our knowledge, we are the first to explicitly formulate the robust real-world deployment of unified VLA policies under sensor-configuration heterogeneity and missing auxiliary sensors as a joint architecture-and-training problem. CoRE-VLA addresses this formulation by treating auxiliary-sensor availability as an action-generator-level routing condition through context-conditioned sparse computation.
\item We enable RGB-pretrained VLA policies to be extended with auxiliary perceptual modalities through continued training, while preserving reliable execution when auxiliary sensors are unavailable, via modality-specialized experts, modality dropout, and availability-based expert masking.
\item We demonstrate strong performance on LIBERO, RoboCasa GR1 Tabletop, and real-world dual-arm manipulation, showing improved multi-task performance, long-horizon execution, missing-depth robustness, and zero-shot generalization to unseen manipulation scenarios.

\end{itemize}

\begin{figure}[t]
    \centering
    \includegraphics[width=\linewidth]{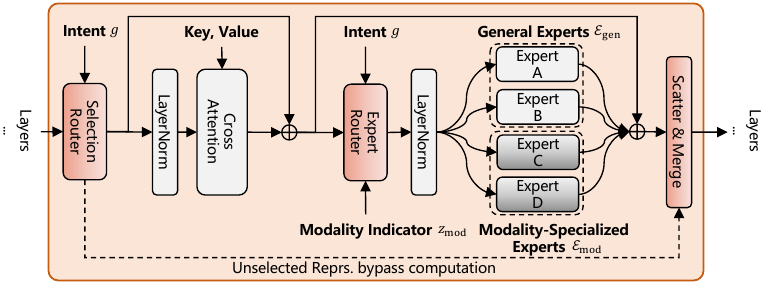}
    \vspace{-15pt}
    \caption{Details of the CoRE block in the Action DiT. Guided by task intent, CoRE first sparsely selects the most task-relevant action-side representations, and then routes them to general or modality-specialized experts conditioned on task intent and the modality indicator.}
    \vspace{-5pt}
    \label{fig:details}
\end{figure}

\section{Related Work}
\label{sec:related_work}

\subsection{Vision-Language-Action Models for Robotic Manipulation}
Vision-language-action (VLA) models unify visual perception, language understanding, and action generation for language-conditioned robotic manipulation. Representative systems, including RT-1~\cite{brohan2022rt}, RT-2~\cite{zitkovich2023rt}, Octo~\cite{team2024octo}, OpenVLA~\cite{kim2025openvla}, $\pi_0$~\cite{black2024pi_0}, and GR00T N1~\cite{bjorck2025gr00t}, demonstrate the scalability of generalist robot policies trained with diverse robot data and pretrained vision-language models. However, they typically rely on largely shared dense action generators across tasks and sensory contexts, which can limit scalability when handling heterogeneous tasks, long-horizon behaviors, and extension to auxiliary sensors.

\subsection{Auxiliary Sensors and Modality-Robust Robot Learning}
Auxiliary physical sensors provide important cues for robotic manipulation beyond RGB observations. Prior work has shown that depth sensing, tactile sensing, and force/torque feedback can improve spatial reasoning, contact-rich manipulation, and fine-grained force control~\cite{shridhar2023perceiver,goyal2023rvt,chen2023visuo,yu2026forcevla}. However, these sensors are not always available across robot platforms and can fail unexpectedly during operation. Policies that tightly couple action generation to auxiliary sensor inputs may therefore over-rely on them and become brittle when the sensors are missing. CoRE-VLA addresses this issue by treating auxiliary sensors as optional conditioning signals: it exploits auxiliary cues when they are available, while preserving reliable action generation when they are absent.


\subsection{Conditional Computation for Scalable Action Generation}
Sparse conditional computation in transformer architectures increases effective capacity by activating only part of the computation for each input, as studied in Mixture-of-Experts (MoE) and related conditional-computation methods~\cite{jacobs1991adaptive,shazeer2017outrageously,lepikhin2020gshard,fedus2022switch,raposo2024mixture}.
Such mechanisms are also relevant to multi-task learning, where forcing heterogeneous objectives to share the same parameters can lead to gradient conflicts, task interference, and negative transfer~\cite{yu2020gradient,standley2020tasks}.
Recent VLA works have explored MoE-style routing to decouple reasoning from control or to preserve pretrained reasoning capabilities during policy learning~\cite{zhou2025chatvla,zhou2026chatvla}.
In contrast, CoRE-VLA is designed to condition action generation on both task intent and sensor availability, enabling task- and modality-adaptive computation for diverse tasks, long-horizon subgoals, and sensor configurations.

\section{Method}
\label{sec:method}

\subsection{Conditional Routing of Experts}

\paragraph{Intent-conditioned routing.}
In VLA models, visual tokens typically substantially outnumber text tokens in the multimodal representation, causing pooled multimodal features to be dominated by visual information and potentially diluting the task intent expressed by the instruction.
We therefore explicitly decouple the routing signal from the visual-token-dominated multimodal representation by pooling only the VLM-encoded text tokens and using the resulting task-intent embedding $g$ to directly guide expert routing in the MoE action head:
\begin{equation}
    g
    =
    \operatorname{AvgPool}(C_{\mathrm{text}})
    \in
    \mathbb{R}^{d_c},
    \label{eq:text_intent_pooling}
\end{equation}
where $C_{\mathrm{text}}$ denotes the VLM-encoded text representations.
Although $g$ is pooled from text tokens, these representations have already interacted with visual tokens through the VLM and are therefore grounded in the current visual context.
This yields an instruction-centered yet visually grounded intent representation for routing action-side representations to task-relevant experts.

\paragraph{Action-side representation selection.}
Real-time closed-loop control requires VLA policies to allocate action-generation computation efficiently.
Instead of applying cross-attention and expert computation uniformly to all action-side representations, CoRE uses task intent to sparsely select a capacity-limited subset of the most task-relevant representations for conditional computation in each CoRE block.
Given the task-intent embedding $g$, each action-side representation $h_i^\ell\in\mathbb{R}^{d}$ in the sequence $H^\ell=\{h_i^\ell\}_{i=1}^{L_a}$ is assigned a selection score $p_i$:
\begin{equation}
    p_i
    =
    \sigma
    \left(
        w_{\mathrm{sel}}^\top [h_i^\ell;g]
    \right),
    \qquad
    \mathcal{S}
    =
    \operatorname{TopK}
    \left(
        \{p_i\}_{i=1}^{L_a},
        \max(1,\lfloor\rho L_a\rfloor)
    \right).
\end{equation}
where $w_{\mathrm{sel}}\in\mathbb{R}^{d+d_c}$ is a learnable selection vector, and $\rho$ controls the activation capacity.

As shown in Figure~\ref{fig:details}, only representations in $\mathcal{S}$ undergo cross-attention and expert routing while unselected representations bypass these computations.
The updated selected representations are then scattered and merged with the bypassed representations to recover the full action-side sequence.

\paragraph{General and modality-specialized experts.}


To support scalable and robust extension to auxiliary perceptual modalities, we divide the expert set into general experts~$\mathcal{E}_{\mathrm{gen}}$ and modality-specialized experts~$\mathcal{E}_{\mathrm{mod}}$, as shown in Figure~\ref{fig:details}.
General experts capture shared and reusable manipulation patterns, while modality-specialized experts handle modality-dependent computation.
This design allows an RGB-pretrained VLA policy to be extended with auxiliary modality encoders and modality-specialized experts through continued training, without redesigning the entire action backbone or retraining it from scratch.

\paragraph{Modality dropout and modality-specialized expert masking.}
To prevent the policy from over-relying on auxiliary sensor inputs, we introduce a modality indicator $z_{\mathrm{mod}}$ for modality dropout and availability-based expert masking.
During training, $z_{\mathrm{mod}}$ randomly disables the auxiliary modality branch and masks modality-specialized experts, exposing the policy to both full-modality and RGB-language-proprioception regimes.
This encourages the policy to exploit auxiliary perceptual signals when they are available while preserving reliable action generation when auxiliary sensors are unavailable during deployment.
Specifically, we sample $z_{\mathrm{mod}}$ with modality dropout probability $p_{\mathrm{drop}}$:
\begin{equation}
    z_{\mathrm{mod}} \sim \operatorname{Bernoulli}(1-p_{\mathrm{drop}}),
    \label{eq:modality_dropout}
\end{equation}
where $z_{\mathrm{mod}}\in\{0,1\}$ is an auxiliary-modality availability indicator.
When $z_{\mathrm{mod}}=1$, the auxiliary modality branch is enabled and modality-specialized experts are available for routing.
When $z_{\mathrm{mod}}=0$, the auxiliary modality branch is disabled and modality-specialized experts are masked.
During inference, $z_{\mathrm{mod}}$ is set according to the availability of the auxiliary sensor on the deployed robot.

For each selected representation $h_i^\ell$ in the current CoRE block, $h_i^\ell$ attends to the condition sequence $C$, producing a conditioned representation $\tilde{h_i^\ell}$.
When $z_{\mathrm{mod}}=0$, auxiliary modality representations are excluded from $C$.
CoRE then predicts expert-routing logits from the conditioned representation and the task-intent embedding:
\begin{equation}
    r_{i,e}
    =
    w_e^\top[\tilde{h_i^\ell};g],
\end{equation}
where $w_e$ is a learnable routing vector for expert $e$.
Before expert selection, we apply an availability-based expert mask to the routing logits:
\begin{equation}
    \bar{r}_{i,e}
    =
    \begin{cases}
        r_{i,e}, 
        & e\in\mathcal{E}_{\mathrm{gen}} \ \text{or}\ 
          \left(e\in\mathcal{E}_{\mathrm{mod}} \ \text{and}\ z_{\mathrm{mod}}=1\right),\\
        -\infty,        
        & e\in\mathcal{E}_{\mathrm{mod}} \ \text{and}\ z_{\mathrm{mod}}=0.
    \end{cases}
    \label{eq:masked_expert_routing}
\end{equation}
For each selected representation, the selected expert and routing probability are
\begin{equation}
    e_i
    =
    \arg\max_e
    \bar{r}_{i,e},
    \qquad
    \pi_i
    =
    \operatorname{softmax}
    (\bar{r}_{i,:})_{e_i}.
    \label{eq:expert_selection}
\end{equation}

\paragraph{Differentiable routing path.}
Although Top-$K$ representation selection and hard expert assignment are discrete and treated as fixed during backpropagation, the selected computation path remains differentiable because the continuous scores $p_i$ and $\pi_i$ are multiplied with the CoRE update and routed expert output, respectively.
Therefore, the representation selector and expert router receive gradients through these continuous gating scores along the selected computation path.

\subsection{Training Objective and Inference}
\label{sec:training_and_inference}

CoRE-VLA is trained with a flow matching objective over action chunks.
Given ground-truth action chunk $\mathbf{a}=\mathbf{a}_{t:t+H}$, proprioceptive state $s_t$, we sample $\boldsymbol{\epsilon}\sim\mathcal{N}(0,I)$ and $\tau\sim\mathcal{U}(0,1)$, and construct
\begin{equation}
    \mathbf{x}_\tau
    =
    (1-\tau)\boldsymbol{\epsilon}
    +
    \tau\mathbf{a},
    \qquad
    \mathbf{v}
    =
    \mathbf{a}
    -
    \boldsymbol{\epsilon}.
    \label{eq:flow_matching_interpolation}
\end{equation}
The model predicts $\hat{\mathbf{v}}_\theta=
f_{\theta,\mathcal{R}(g,z_{\mathrm{mod}})}(\mathbf{x}_\tau,\tau,s_t,C)$ and is optimized with
\begin{equation}
    \mathcal{L}_{\mathrm{act}}
    =
    \mathbb{E}
    \left[
        \|\hat{\mathbf{v}}_\theta-\mathbf{v}\|_2^2
    \right].
    \label{eq:flow_matching_loss}
\end{equation}
To stabilize the gradient path through the continuous representation-selection scores, we use a selection regularizer that aligns the average representation-selection mass with the target capacity ratio $\rho$:
\begin{equation}
    \mathcal{L}_{\mathrm{sel}}
    =
    \sum_{\ell\in\mathcal{I}_{\mathrm{CoRE}}}
    \left(
        \frac{1}{L_a}
        \sum_{i=1}^{L_a}
        p_i^\ell
        -
        \rho
    \right)^2,
    \label{eq:selection_regularization}
\end{equation}
where $\mathcal{I}_{\mathrm{CoRE}}$ denotes the set of layers equipped with CoRE blocks.

The standard MoE load-balancing loss $\mathcal{L}_{\mathrm{moe}}$ is applied over selected representations and available experts to encourage balanced expert utilization and prevent routing collapse~\cite{fedus2022switch}.

The final training objective is:
\begin{equation}
    \mathcal{L}
    =
    \mathcal{L}_{\mathrm{act}}
    +
    \lambda_{\mathrm{sel}}\mathcal{L}_{\mathrm{sel}}
    +
    \lambda_{\mathrm{moe}}\mathcal{L}_{\mathrm{moe}},
    \label{eq:total_training_loss}
\end{equation}
where $\lambda_{\mathrm{sel}}$ and $\lambda_{\mathrm{moe}}$ are weighting coefficients for $\mathcal{L}_{\mathrm{sel}}$ and $\mathcal{L}_{\mathrm{moe}}$, respectively.

At inference time, CoRE-VLA generates an action chunk by integrating the learned velocity field from Gaussian noise to the action space. Specifically, we sample $\mathbf{x}_0\sim\mathcal{N}(0,I)$ and solve
\begin{equation}
    \frac{d\mathbf{x}_\tau}{d\tau}
    =
    f_{\theta,\mathcal{R}(g,z_{\mathrm{mod}})}(\mathbf{x}_\tau,\tau,s_t,C),
    \qquad
    \tau:0\rightarrow 1.
    \label{eq:inference_ode}
\end{equation}
The final state $\mathbf{x}_1$ is used as the predicted action chunk $\hat{\mathbf{a}}_{t:t+H}$ for closed-loop execution.



\section{Experimental Results}
\label{sec:result}

\subsection{Simulation Results}

We first evaluate CoRE-VLA on two simulation benchmarks: LIBERO~\cite{liu2023libero} and RoboCasa GR1 Tabletop~\cite{nasiriany2024robocasa,bjorck2025gr00t}.
In simulation, we instantiate the auxiliary perceptual modality as depth to evaluate both modality extensibility and robustness to missing auxiliary inputs.
Instead of using overly clean ground-truth simulation depth, we estimate depth maps from RGB observations with Depth Anything V2 (DA-V2)~\cite{yang2024depthv2} and use them as noisy auxiliary depth inputs during training and evaluation.
For each benchmark, we jointly train all tasks with a single CoRE-VLA policy on 8 NVIDIA H200 GPUs, using Qwen3-VL-4B-Instruct~\cite{bai2025qwen3} as the VLM backbone.
We first train a dense Action DiT from scratch on RGB-language-proprioception inputs, then convert it into CoRE-VLA by duplicating FFN modules as general experts, inserting modality-specialized experts, and adding an auxiliary sensor encoder, followed by continued training with auxiliary perceptual inputs.
Additional experimental details are provided in Appendix~\ref{sec:details_exp}.

\begin{wraptable}{r}{0.55\textwidth}
\footnotesize
\setlength{\tabcolsep}{1.7pt}
\renewcommand{\arraystretch}{0.85}
\vspace{-15pt}
\caption{Results on the LIBERO benchmark.}
\label{tab:libero}
\centering
\begin{tabular}{lccccc}
\toprule
\textbf{Method} & \textbf{Spatial} & \textbf{Object} & \textbf{Goal} & \textbf{Long} & \textbf{Average} \\
\midrule
OpenVLA~\cite{kim2025openvla} & 84.7 & 88.4 & 79.2 & 53.7 & 76.5 \\
OpenVLA-OFT~\cite{kim2025fine} & 97.6 & 98.4 & 97.9 & 94.5 & 97.1 \\
$\pi_0$~\cite{black2024pi_0} & 96.8 & 98.8 & 95.8 & 85.2 & 94.2 \\
$\pi_0$+FAST~\cite{black2024pi_0,pertsch2025fast} & 96.4 & 96.8 & 88.6 & 60.2 & 85.5 \\
$\pi_{0.5}$~\cite{intelligence2025pi_} & 98.8 & 98.2 & 98.0 & 92.4 & 96.9 \\
GR00T-N1.5~\cite{bjorck2025gr00t,nvidia2025isaacgr00tn15libero} & 92.0 & 92.0 & 86.0 & 76.0 & 86.5 \\
GR00T-N1.7~\cite{bjorck2025gr00t,nvidia2026isaacgr00tn17libero} & 97.7 & 98.5 & 97.5 & 94.4 & 97.0 \\
\midrule
CoRE-VLA (Ours) & \textbf{99.0} & \textbf{99.2} & \textbf{98.8} & \textbf{97.6} & \textbf{98.7} \\
\bottomrule
\end{tabular}
\vspace{-10pt}
\end{wraptable}

\vspace{-5pt}
\paragraph{LIBERO benchmark results.}
LIBERO covers four representative manipulation suites: spatial reasoning, object-centric manipulation, goal-conditioned manipulation, and long-horizon task execution.
Each suite contains 10 tasks.
We evaluate CoRE-VLA with 50 rollouts per task.
As shown in Table~\ref{tab:libero}, CoRE-VLA achieves the best performance
across all four suites, with the largest advantage on LIBERO-Long.
We therefore use LIBERO-Long for the controlled analyses below.
Detailed component-wise ablations and routing visualizations are provided in
Appendices~\ref{sec:ablation_analysis} and~\ref{sec:router_analysis}.
\vspace{-5pt}
\begin{table}[H]
\centering
\footnotesize
\caption{Controlled LIBERO-Long comparisons.}
\vspace{-5pt}
\label{tab:controlled}

\centering
\setlength{\tabcolsep}{35pt}
\renewcommand{\arraystretch}{0.8}
\begin{tabular}{@{}lcc@{}}
\toprule
Model / Variant & \dplus{} SR & \dminus{} SR \\
\midrule
Dense RGB (RGB-only) & -- & 94.4 \\
Dense RGB-D (no modality dropout) & 94.6 & 72.0 \\
Dense RGB-D + modality dropout & 94.8 & 94.8 \\
Dense RGB-D + modality dropout (parameter-matched) & 95.2 & 95.0 \\
Dense RGB-D + modality dropout (FLOPs-matched) & 93.4 & 93.4 \\
MoE RGB-D + modality dropout (fixed expert routing) & 94.6 & 94.6 \\
CoRE-VLA (intent + modality conditional routing) & 97.6 & 97.0 \\
\bottomrule
\end{tabular}

\vspace{2pt}
{\scriptsize
\raggedright
\dplus/\dminus{} denote depth enabled/disabled at inference.
\par}

\vspace{-15pt}
\end{table}
\paragraph{Controlled analysis on LIBERO-Long.}
To disentangle the effects of auxiliary depth, model capacity, and conditional routing,
we conduct controlled comparisons on LIBERO-Long in Table~\ref{tab:controlled}.
A dense RGB-D policy without modality dropout substantially degrades when depth is
removed ($94.6\%\rightarrow72.0\%$), while modality dropout restores missing-depth
robustness. Importantly, parameter-matched and FLOPs-matched dense baselines, as well
as a fixed-routing MoE, remain below CoRE-VLA. CoRE-VLA achieves $97.6\%$ with
depth and $97.0\%$ without depth, showing that its gains cannot be explained by
additional depth input or model capacity alone, but benefit from learned conditional routing.
Additional component-wise ablations are provided in Appendix~\ref{sec:ablation_analysis}.

\vspace{-5pt}
\paragraph{RoboCasa GR1 Tabletop benchmark results.}
The RoboCasa GR1 Tabletop benchmark contains $24$ tabletop manipulation tasks with teleoperated demonstrations and evaluates generalist robotic policies in closed-loop simulation.
We train CoRE-VLA with $1000$ demonstrations per task and evaluate it with $5$ independent groups of $100$ rollouts.
As reported in Table~\ref{tab:robocasa_gr1}, CoRE-VLA achieves the best average success rate, validating its multi-task manipulation capability across diverse tabletop scenarios.
Compared with dense action generators that share the same computation path across all tasks, CoRE-VLA introduces conditional routing within the action backbone, enabling task-dependent expert specialization while preserving shared general manipulation knowledge.
This design can mitigate negative transfer and optimization interference among heterogeneous tasks, leading to better scalability and robustness in multi-task policy learning.

\begin{table}[H]
\footnotesize
\caption{Results on the RoboCasa GR1 Tabletop benchmark. CoRE-VLA is reported as mean $\pm$ std over $5$ independent groups of $100$ rollouts. See Appendix~\ref{sec:detial_simulation} for details.}
\vspace{-5pt}
\label{tab:robocasa_gr1}
\centering
\setlength{\tabcolsep}{3pt}
\renewcommand{\arraystretch}{0.85}
\begin{tabular}{lcccc}
\toprule
\textbf{Method} 
& \textbf{Diffusion Policy}~\cite{chi2025diffusion,bjorck2025gr00t}
& \textbf{GR00T-N1.5}~\cite{bjorck2025gr00t,nvidia2025isaacgr00tn15robocasa} 
& \textbf{GR00T-N1.6}~\cite{bjorck2025gr00t,nvidia2026isaacgr00tn16robocasa} 
& \textbf{CoRE-VLA} \\
\midrule
\textbf{Average Success Rate} & 40.4 & 48.0 & 47.6 & $\mathbf{56.5\pm0.4}$ \\

\bottomrule
\end{tabular}
\vspace{2pt}
\end{table}

\begin{figure}[h!]
    \centering
    \includegraphics[width=1\linewidth]{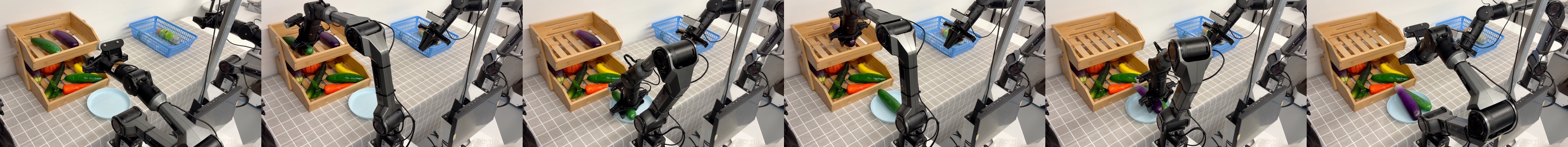}
    \vspace{-20pt}
    \caption{Temporal snapshots of CoRE-VLA real-world rollouts on the \textbf{Vegetables-Picking} task. The robot picks up the cucumber and eggplant from the upper layer of the fruit basket and places them on the plate. This task evaluates the model’s spatial understanding, target-location generalization, object grounding, and long-horizon manipulation capability.}
    \label{fig:pick_vegetables}
\end{figure}

\begin{figure}[h!]
\vspace{-10pt}
    \centering
    \includegraphics[width=1\linewidth]{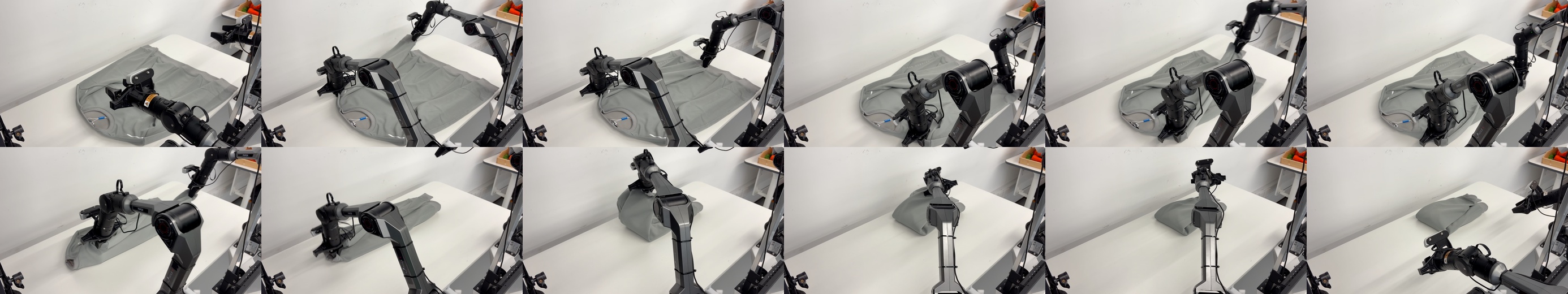}
    \vspace{-20pt}
    \caption{Temporal snapshots of CoRE-VLA real-world rollouts on the \textbf{Clothes-Folding} task. The robot uses both arms to fold the left and right sleeves followed by two garment-folding stages. This task evaluates the model’s ability to perform complex bimanual manipulation of deformable objects and long-horizon contact-rich manipulation.}
    \label{fig:fold_sleeve}
\end{figure}

\begin{figure}[h!]
\vspace{-10pt}
    \centering
    \includegraphics[width=1\linewidth]{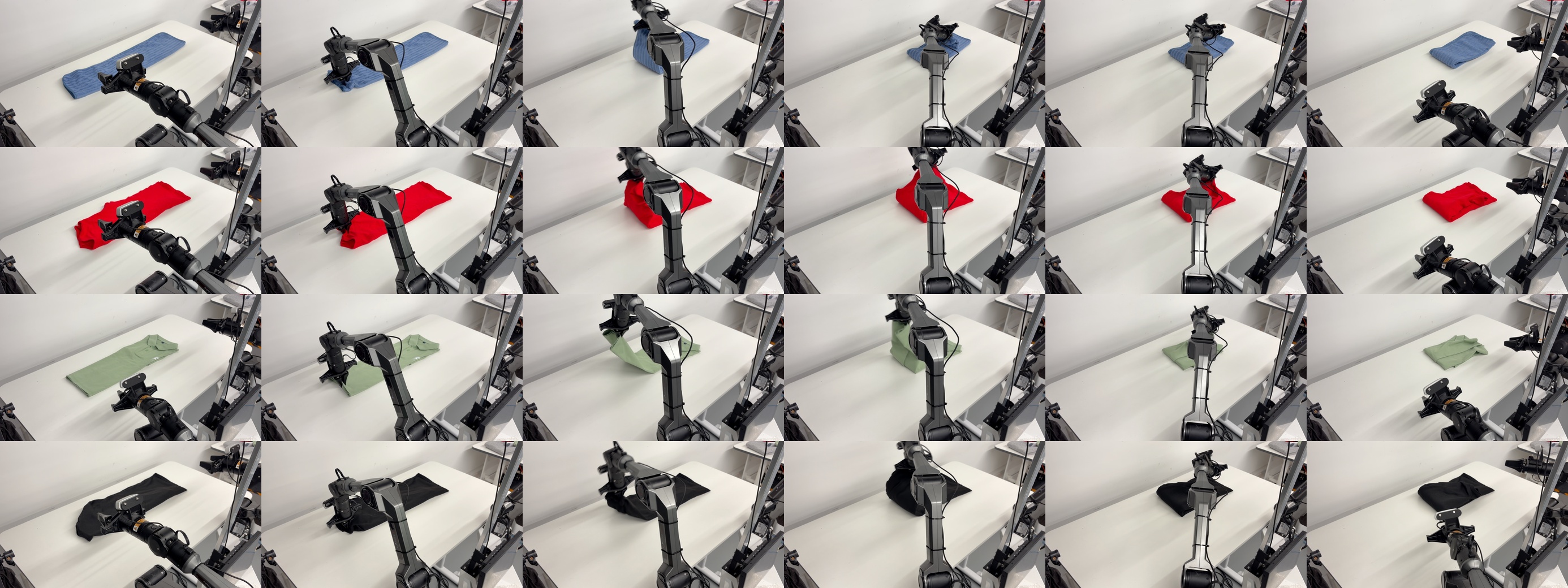}
    \vspace{-20pt}
    \caption{Temporal snapshots of CoRE-VLA real-world rollouts on the \textbf{Fabric-Folding} task. Although trained only on gray clothes-folding demonstrations, the model zero-shot generalizes to fabric-folding scenarios with diverse colors, shapes, and materials.}
    \label{fig:fold_fabric}
\end{figure}

\subsection{Real-World Results}

\paragraph{Real-world manipulation benchmark.}
We further evaluate CoRE-VLA on a challenging real-world manipulation benchmark consisting of three tasks: Vegetables-Picking, Clothes-Folding, and Fabric-Folding. The benchmark is designed to assess spatial understanding, target-location generalization, object grounding, complex bimanual manipulation of deformable objects, long-horizon contact-rich manipulation, and zero-shot generalization. In addition, we evaluate the RGB-D-trained CoRE-VLA model under two inference settings: without auxiliary depth and with a physical depth camera, to validate the modality robustness of CoRE-VLA in the physical world.
\vspace{-10pt}
\paragraph{Robot platform and data collection.}
All real-world experiments are conducted on an AgileX ALOHA dual-arm robot with one head-mounted camera and two wrist-mounted cameras.
The policy receives three-view RGB observations, proprioceptive states, and language instructions, and predicts action chunks for closed-loop execution.
We collect 44 demonstrations for Vegetables-Picking and 1.3K demonstrations for Clothes-Folding through leader-follower teleoperation.
Fabric-Folding is evaluated in a zero-shot setting, where the policy is trained only on Clothes-Folding demonstrations and directly tested on fabrics with unseen colors, shapes, and materials.
Because collecting large-scale demonstrations with synchronized physical depth sensors is costly and hardware-dependent, CoRE-VLA uses DA-V2 pseudo-depth extracted from RGB observations during training.
During deployment, we evaluate the same policy either without auxiliary depth or with a raw physical depth camera, without sensor-specific fine-tuning.

\vspace{-10pt}
\paragraph{Training setup.}
All compared policies are fully fine-tuned on the same real-world training demonstrations using 8 NVIDIA H200 GPUs.
$\pi_{0.5}$ is initialized from its publicly released pretrained JAX version checkpoint before fine-tuning~\cite{intelligence2025pi_}, whereas the baseline and CoRE-VLA use Qwen3-VL-4B-Instruct as the vision-language backbone and train the action generator under the same real-world post-training setup. The baseline uses a dense Action DiT without auxiliary depth modality, while CoRE-VLA uses conditional routing and auxiliary-depth conditioning.
\vspace{-10pt}
\paragraph{Real-world evaluation protocol.}
We report both average task score (Score) and average success rate (SR) in Table~\ref{tab:real-world}.
The task score is computed based on predefined subgoals for each task, and each rollout is scored on a 100-point scale.
The scoring criteria for each task are provided in Appendix~\ref{sec:detail_real_world}.
Average success rate measures the percentage of rollouts that complete all task subgoals.
Each method is evaluated over $20$ independent rollouts per task under the same protocol.

\begin{table}[t!]
\footnotesize
\setlength{\tabcolsep}{3.2pt}
\renewcommand{\arraystretch}{0.85}
\caption{Results on the Real-World Experiments. See Appendix~\ref{sec:detail_real_world} for details.}
\label{tab:real-world}
\vspace{-5pt}
\centering
\begin{tabular}{lcccccc}
\toprule
\multirow{2}{*}[-0.7ex]{\textbf{Method}} 
& \multicolumn{2}{c}{\textbf{Vegetables-Picking}} 
& \multicolumn{2}{c}{\textbf{Clothes-Folding}} 
& \multicolumn{2}{c}{\textbf{Fabric-Folding}} \\
\cmidrule(lr){2-3} \cmidrule(lr){4-5} \cmidrule(lr){6-7}
& \textbf{Score} & \textbf{SR}
& \textbf{Score} & \textbf{SR}
& \textbf{Score} & \textbf{SR} \\
\midrule
$\pi_{0.5}$ (JAX version)~\cite{intelligence2025pi_} 
& 72.5 & 55.0 
& 73.8 & 60.0 
& 62.5 & 60.0 \\
Baseline (Dense Action DiT w/o depth modality)
& 57.5 & 35.0 
& 67.5 & 30.0 
& 62.5 & 50.0 \\
\midrule
CoRE-VLA (Infer. w/o auxiliary depth)
& 77.5 & 65.0
& 77.5 & 50.0
& 75.0 & 70.0 \\
CoRE-VLA (Infer. w/ physical depth camera)
& \textbf{78.8} & \textbf{70.0}
& \textbf{78.8} & \textbf{65.0}
& \textbf{87.5} & \textbf{80.0} \\
\bottomrule
\end{tabular}
\vspace{-10pt}
\end{table}

\vspace{-10pt}
\paragraph{Results and ablation analysis.}
Temporal snapshots of CoRE-VLA real-world rollouts are shown in Figures~\ref{fig:pick_vegetables}, \ref{fig:fold_sleeve}, and \ref{fig:fold_fabric}.
As shown in Table~\ref{tab:real-world}, CoRE-VLA achieves the strongest overall performance across the three real-world tasks.
The two CoRE-VLA inference settings further evaluate robustness under train--deployment auxiliary-modality shifts. Even without auxiliary depth at inference, CoRE-VLA consistently outperforms the dense Action DiT baseline, showing that modality dropout and modality-specialized expert masking prevent over-reliance on depth and preserve reliable policy execution from RGB, language, and proprioceptive inputs alone. When a physical depth camera is enabled, CoRE-VLA further improves performance, indicating that modality-specialized experts can exploit geometric cues when they are available.
The gains are particularly clear on Fabric-Folding, where CoRE-VLA demonstrates strong zero-shot generalization to deformable-object manipulation. CoRE-VLA with physical depth also outperforms the strong pretrained $\pi_{0.5}$ baseline across all three tasks, despite using an action generator trained without large-scale robot policy pretraining. Together, these results show that CoRE-VLA remains robust when auxiliary depth is unavailable while benefiting from physical depth when available, supporting scalable and modality-robust VLA modeling in real-world deployment.

\subsection{Efficiency and Robustness Analysis}

\paragraph{Conditional-computation efficiency.}
As shown in Table~\ref{tab:efficiency}, CoRE-VLA matches the total parameter count
of the parameter-matched dense baseline (388.3M) while activating fewer parameters
(269.5M vs.\ 387.5M), using fewer FLOPs (159.1 vs.\ 235.9 GFLOPs/query), and
reducing end-to-end latency (346.9 vs.\ 555.7 ms).
During real-world execution, the policy replans every 25 actions; at a 50-Hz action
execution rate, its 346.9-ms latency fits within the 500-ms replanning window.
\begin{table}[h]
\centering
\footnotesize
\caption{Efficiency analysis of the action generator and end-to-end policy.}
\vspace{-5pt}
\label{tab:efficiency}
\renewcommand{\arraystretch}{0.85}
\setlength{\tabcolsep}{12pt}
\begin{tabular}{@{}lccccc@{}}
\toprule
Model
& \shortstack{AG Params\\T / A (M)}
& \shortstack{AG\\GFLOPs/q}
& \shortstack{AG Lat.\\(ms)}
& \shortstack{E2E Lat.\\(ms)}
& \shortstack{E2E Mem.\\(GiB)} \\
\midrule
Dense RGB-D
& 161.5 / 160.7
& 163.3
& 90.7
& 531.6
& 9.1 \\

Dense RGB-D (PM)
& 388.3 / 387.5
& 235.9
& 96.3
& 555.7
& 9.5 \\

CoRE-VLA
& 388.3 / 269.5
& 159.1
& 77.2
& 346.9
& 9.5 \\
\bottomrule
\end{tabular}

\vspace{2pt}
{\scriptsize
AG: Action Generator; PM: parameter-matched; T/A: total/activated parameters;
GFLOPs/q: GFLOPs per policy query; E2E: end-to-end; Lat.: latency;
Mem.: peak allocated GPU memory.
}
\vspace{-10pt}
\end{table}

\paragraph{Robustness to depth degradation.}
We further evaluate CoRE-VLA on LIBERO-Long under realistic depth degradations, including mid-rollout outages, intermittent loss, noise, delay, and spatial misalignment.
CoRE-VLA remains stable at $97.0$--$97.6\%$ success across all tested conditions, whereas the dense RGB-D baseline drops to $72.0\%$ under complete removal and $86.8\%$ under intermittent loss.
These results demonstrate robustness beyond complete auxiliary-depth absence.
\begin{table}[h]
\centering
\footnotesize
\caption{Sensor-failure robustness on LIBERO-Long under different depth degradations.}
\vspace{-5pt}
\label{tab:failure}
\renewcommand{\arraystretch}{0.85}
\setlength{\tabcolsep}{24pt}
\begin{tabular}{@{}lccc@{}}
\toprule
Depth condition
& $z_{\rm mod}$
& Dense RGB-D
& CoRE-VLA \\
\midrule
Available throughout
& 1
& 94.6
& 97.6 \\

Unavailable throughout
& 0
& 72.0
& 97.0 \\

Mid-rollout outage (50\% horizon)
& $1\!\rightarrow\!0$
& 93.0
& 97.2 \\

Intermittent loss, $p=0.5$
& $1/0$
& 86.8
& 97.4 \\

Gaussian noise: $\sigma=0.1\operatorname{mean}|d|$
& 1
& 94.0
& 97.4 \\

One policy-query interval delay
& 1
& 93.8
& 97.4 \\

Right shift (16 px, zero-padded)
& 1
& 93.6
& 97.4 \\
\bottomrule
\end{tabular}
\vspace{-10pt}
\end{table}

\section{Limitations}

While CoRE-VLA is designed to accommodate different auxiliary perceptual modalities, 
our current empirical validation focuses on optional depth input rather than arbitrary 
multi-sensor configurations. Evaluation with richer modalities such as tactile sensing 
and force feedback, larger sensor suites, and broader cross-embodiment settings remains 
future work.




\section{Conclusion}
\label{sec:conclusion}
We presented CoRE-VLA, a vision-language-action framework that formulates action generation as context-conditioned sparse computation. 
To preserve task intent from visual-token-dominated multimodal representations, CoRE-VLA pools VLM-encoded text representations as an explicit intent signal to guide expert routing in the action head, while sensor availability gates modality-specialized experts. 
Together with modality dropout, this design enables the policy to adapt its computation to both task intent and available sensory context.

Across LIBERO, RoboCasa GR1 Tabletop, and real-world dual-arm manipulation, CoRE-VLA achieves strong performance on multi-task and long-horizon manipulation. 
Our ablations show that text-intent-conditioned routing outperforms routing based on pooled full vision-language representations, while modality analysis demonstrates that CoRE-VLA can exploit auxiliary depth when available and remain robust when depth is unavailable at inference time. 
These results suggest that effective VLA action generation benefits from explicitly preserving task intent for conditional computation while maintaining robustness to changes in auxiliary sensory inputs.

\clearpage

\bibliography{example}  

\clearpage

\appendix

\section{Details of Experiments}
\label{sec:details_exp}

\subsection{Training Configurations}
We provide detailed implementation and reproducibility information for LIBERO and RoboCasa GR1 in Tables~\ref{tab:libero_reproducibility_details} and~\ref{tab:gr1_reproducibility_details}. 
These tables summarize the dataset setup, model architecture, training hyperparameters, routing configuration, and evaluation protocol used in our experiments.

\begin{table}[h]
\centering
\caption{Implementation and reproducibility details for LIBERO experiments.}
\label{tab:libero_reproducibility_details}
\small
\setlength{\tabcolsep}{6pt}
\begin{tabular}{ll}
\toprule
\textbf{Item} & \textbf{Value} \\
\midrule
Dataset & LIBERO 4 suites \\
Data mix & \texttt{libero\_all} \\
Policy input & RGB, proprioception, language, depth \\
Image size & $224 \times 224$ \\
VLM backbone & Qwen3-VL-4B-Instruct \\
Action model & DiT-B with CoRE routing \\
Action DiT layers & 16 \\
Hidden size & 1024 \\
Cross-attention dim & 2560 \\
Action dim & 7 \\
State dim & 7 \\
Action horizon & 8 \\
Future action window size & 7 \\
Training steps & 100K \\
Effective batch size & 128 \\
GPUs & 8$\times$H200 \\
Optimizer & AdamW \\
Warmup & 5K steps \\
LR schedule & cosine with minimum LR $1\times 10^{-6}$ \\
Action model LR & $1\times 10^{-4}$ \\
VLM/interface LR & $1\times 10^{-5}$ \\
Base LR & $2.5\times 10^{-5}$ \\
Gradient clipping & 1.0 \\
Weight decay & $1\times 10^{-8}$ \\
Fine-tuning strategy & Full fine-tuning \\
Auxiliary modality encoder & Conv Projector (kernel=$28\times28$, stride=$28$) \\
Auxiliary modality dropout probability & 0.2 \\
Total experts & 12 \\
General experts & 8 \\
Modality-specialized experts & 4 \\
Routing mode & Token-choice routing \\
CoRE layers & Layers 2, 4, 6, and 8 \\
Selection-capacity ratio & 0.5 \\
MoE balancing loss weight & 0.01 \\
Selection regularization weight & 0.01 \\
Inference diffusion steps & 4 \\
Checkpoint selection & Final checkpoint at 100K steps \\
Eval rollouts & $50$ rollouts per task \\
Seed & 42 \\
\bottomrule
\end{tabular}
\end{table}

\begin{table}[h]
\centering
\caption{Implementation and reproducibility details for RoboCasa GR1 Tabletop experiments.}
\label{tab:gr1_reproducibility_details}
\small
\setlength{\tabcolsep}{6pt}
\begin{tabular}{ll}
\toprule
\textbf{Item} & \textbf{Value} \\
\midrule
Dataset & RoboCasa GR1 / GR00T-X Embodiment Sim \\
Data mix & \texttt{fourier\_gr1\_unified\_1000} \\
Policy input & RGB, proprioception, language, depth \\
Image size & $224 \times 224$ \\
VLM backbone & Qwen3-VL-4B-Instruct \\
Action model & DiT-B with CoRE routing \\
Action DiT layers & 32 \\
Hidden size & 2560 \\
Cross-attention dim & 2560 \\
Action dim & 29 \\
State dim & 58 \\
Action horizon & 16 \\
Future action window size & 15 \\
Training steps & 200K \\
Effective batch size & 128 \\
GPUs & 8$\times$H200 \\
Optimizer & AdamW \\
Warmup & 5K steps \\
LR schedule & cosine with minimum LR $5\times 10^{-7}$ \\
Action model LR & $1\times 10^{-4}$ \\
VLM/interface LR & $1\times 10^{-5}$ \\
Base LR & $4\times 10^{-5}$ \\
Gradient clipping & 1.0 \\
Weight decay & $1\times 10^{-8}$ \\
Fine-tuning strategy & Full fine-tuning \\
Auxiliary modality encoder & Conv Projector (kernel=$28\times28$, stride=$28$) \\
Auxiliary modality dropout probability & 0.2 \\
Total experts & 32 \\
General experts & 24 \\
Modality-specialized experts & 8 \\
Routing mode & Token-choice routing \\
CoRE layers & Layers 8, 10, 12, 14, 16, 18, 20, and 22 \\
Selection-capacity ratio & 0.5 \\
MoE balancing loss weight & 0.01 \\
Selection regularization weight & 0.01 \\
Inference diffusion steps & 4 \\
Checkpoint selection & Final checkpoint at 200K steps \\
Eval rollouts & $5 \times 100$ rollouts per task \\
Seed & 42 \\
\bottomrule
\end{tabular}
\end{table}

\clearpage

\subsection{Details of Simulation Evaluation}
\label{sec:detial_simulation}

We provide detailed quantitative results to complement the aggregate performance reported in the main paper. 
For simulation experiments, we include group-wise evaluation statistics on RoboCasa GR1 Tabletop, detailed per-task comparisons with representative baselines, and per-task success rates on the four LIBERO suites. 
For real-world experiments, we report the score and success indicator of every rollout, rather than only aggregate averages, to make the evaluation protocol and failure distribution transparent. 

\paragraph{RoboCasa GR1 Tabletop.}
Tables~\ref{tab:robocasa_gr1_groupwise} and~\ref{tab:robocasa_gr1_detail} provide detailed results on the RoboCasa GR1 Tabletop benchmark. 
Table~\ref{tab:robocasa_gr1_groupwise} reports the five independent evaluation groups of CoRE-VLA, where each group contains 100 rollouts per task. 
This allows us to measure both the average success rate and the variation across repeated evaluation groups. 
Table~\ref{tab:robocasa_gr1_detail} further compares CoRE-VLA with Diffusion Policy, GR00T-N1.5, and GR00T-N1.6 on each individual task. 
The per-task breakdown shows that CoRE-VLA achieves strong average performance while maintaining competitive results across diverse object, receptacle, and placement configurations.


\begin{table*}[!htbp]
\centering
\footnotesize
\caption{Group-wise results of CoRE-VLA on the RoboCasa GR1 Tabletop benchmark.
Each trial contains 100 rollouts per task.
Each task row reports mean $\pm$ std across the five trials.
For the average row, we first average success rates across all tasks within each trial, and then report mean $\pm$ std over the five trial-level averages.}
\label{tab:robocasa_gr1_groupwise}
\setlength{\tabcolsep}{4pt}
\begin{tabular}{lcccccc}
\toprule
\textbf{Task} 
& \textbf{Trial 1}
& \textbf{Trial 2}
& \textbf{Trial 3}
& \textbf{Trial 4}
& \textbf{Trial 5}
& \textbf{Mean $\pm$ Std} \\
\midrule
PnPBottleToCabinetClose & 78 & 66 & 70 & 74 & 65 & $70.6 \pm 5.5$ \\
PnPCanToDrawerClose & 77 & 72 & 76 & 73 & 67 & $73.0 \pm 3.9$ \\
PnPCupToDrawerClose & 49 & 54 & 40 & 42 & 57 & $48.4 \pm 7.4$ \\
PnPMilkToMicrowaveClose & 51 & 64 & 56 & 50 & 60 & $56.2 \pm 5.9$ \\
PnPPotatoToMicrowaveClose & 42 & 42 & 40 & 28 & 38 & $38.0 \pm 5.8$ \\
PnPWineToCabinetClose & 48 & 52 & 50 & 60 & 45 & $51.0 \pm 5.7$ \\
\addlinespace
PnPNovelFromCuttingboardToBasket & 57 & 55 & 66 & 56 & 57 & $58.2 \pm 4.4$ \\
PnPNovelFromCuttingboardToCardboardbox & 53 & 54 & 51 & 54 & 60 & $54.4 \pm 3.4$ \\
PnPNovelFromCuttingboardToPan & 73 & 70 & 70 & 71 & 71 & $71.0 \pm 1.2$ \\
PnPNovelFromCuttingboardToPot & 58 & 67 & 56 & 64 & 64 & $61.8 \pm 4.6$ \\
PnPNovelFromCuttingboardToTieredbasket & 57 & 54 & 52 & 54 & 51 & $53.6 \pm 2.3$ \\
\addlinespace
PnPNovelFromPlacematToBasket & 50 & 51 & 59 & 60 & 52 & $54.4 \pm 4.7$ \\
PnPNovelFromPlacematToBowl & 60 & 60 & 55 & 66 & 56 & $59.4 \pm 4.3$ \\
PnPNovelFromPlacematToPlate & 64 & 74 & 65 & 60 & 69 & $66.4 \pm 5.3$ \\
PnPNovelFromPlacematToTieredshelf & 26 & 26 & 29 & 32 & 20 & $26.6 \pm 4.4$ \\
\addlinespace
PnPNovelFromPlateToBowl & 67 & 60 & 65 & 60 & 66 & $63.6 \pm 3.4$ \\
PnPNovelFromPlateToCardboardbox & 46 & 49 & 54 & 47 & 54 & $50.0 \pm 3.8$ \\
PnPNovelFromPlateToPan & 53 & 53 & 68 & 60 & 59 & $58.6 \pm 6.2$ \\
PnPNovelFromPlateToPlate & 73 & 86 & 79 & 84 & 74 & $79.2 \pm 5.8$ \\
\addlinespace
PnPNovelFromTrayToCardboardbox & 53 & 55 & 63 & 52 & 54 & $55.4 \pm 4.4$ \\
PnPNovelFromTrayToPlate & 74 & 74 & 66 & 72 & 66 & $70.4 \pm 4.1$ \\
PnPNovelFromTrayToPot & 55 & 57 & 63 & 60 & 58 & $58.6 \pm 3.0$ \\
PnPNovelFromTrayToTieredbasket & 59 & 45 & 47 & 51 & 59 & $52.2 \pm 6.6$ \\
PnPNovelFromTrayToTieredshelf & 28 & 29 & 24 & 21 & 29 & $26.2 \pm 3.6$ \\
\midrule
\textbf{Average Success Rate} & 56.29 & 57.04 & 56.83 & 56.29 & 56.29 & $\mathbf{56.5 \pm 0.4}$ \\
\bottomrule
\end{tabular}
\end{table*}

\clearpage

\begin{table}[!htbp]
\footnotesize
\caption{Detailed per-task comparison on the RoboCasa GR1 Tabletop benchmark. 
For CoRE-VLA, each task is evaluated with $5$ independent groups of $100$ rollouts, and each task row reports mean $\pm$ std across the five groups. 
For the average row, we first compute the average success rate across all tasks within each group, and then report mean $\pm$ std over the five group-level averages.}
\label{tab:robocasa_gr1_detail}
\centering
\setlength{\tabcolsep}{3pt}
\begin{tabular}{lcccc}
\toprule
\textbf{Task} 
& \textbf{Diffusion Policy} 
& \textbf{GR00T-N1.5} 
& \textbf{GR00T-N1.6} 
& \textbf{CoRE-VLA} \\
\midrule
PnPBottleToCabinetClose & 60.8 & 54.0 & 51.5 & $70.6\pm5.5$ \\
PnPCanToDrawerClose & 75.5 & 50.0 & 13.0 & $73.0\pm3.9$ \\
PnPCupToDrawerClose & 36.3 & 38.0 & 8.5 & $48.4\pm7.4$ \\
PnPMilkToMicrowaveClose & 51.0 & 60.0 & 14.0 & $56.2\pm5.9$ \\
PnPPotatoToMicrowaveClose & 41.2 & 32.0 & 41.5 & $38.0\pm5.8$ \\
PnPWineToCabinetClose & 60.8 & 38.0 & 16.5 & $51.0\pm5.7$ \\
\addlinespace
PnPNovelFromCuttingboardToBasket & 29.4 & 38.0 & 58.0 & $58.2\pm4.4$ \\
PnPNovelFromCuttingboardToCardboardbox & 22.6 & 46.0 & 46.5 & $54.4\pm3.4$ \\
PnPNovelFromCuttingboardToPan & 57.8 & 58.0 & 68.5 & $71.0\pm1.2$ \\
PnPNovelFromCuttingboardToPot & 48.0 & 62.0 & 65.0 & $61.8\pm4.6$ \\
PnPNovelFromCuttingboardToTieredbasket & 18.6 & 28.0 & 46.5 & $53.6\pm2.3$ \\
\addlinespace
PnPNovelFromPlacematToBasket & 41.2 & 30.0 & 58.5 & $54.4\pm4.7$ \\
PnPNovelFromPlacematToBowl & 23.5 & 60.0 & 57.5 & $59.4\pm4.3$ \\
PnPNovelFromPlacematToPlate & 37.3 & 56.0 & 63.0 & $66.4\pm5.3$ \\
PnPNovelFromPlacematToTieredshelf & 11.8 & 36.0 & 28.5 & $26.6\pm4.4$ \\
\addlinespace
PnPNovelFromPlateToBowl & 31.4 & 58.0 & 57.0 & $63.6\pm3.4$ \\
PnPNovelFromPlateToCardboardbox & 27.5 & 44.0 & 43.5 & $50.0\pm3.8$ \\
PnPNovelFromPlateToPan & 35.3 & 60.0 & 51.0 & $58.6\pm6.2$ \\
PnPNovelFromPlateToPlate & 61.8 & 64.0 & 78.7 & $79.2\pm5.8$ \\
\addlinespace
PnPNovelFromTrayToCardboardbox & 40.2 & 52.0 & 51.5 & $55.4\pm4.4$ \\
PnPNovelFromTrayToPlate & 49.0 & 48.0 & 71.0 & $70.4\pm4.1$ \\
PnPNovelFromTrayToPot & 52.9 & 60.0 & 64.5 & $58.6\pm3.0$ \\
PnPNovelFromTrayToTieredbasket & 39.2 & 52.0 & 57.0 & $52.2\pm6.6$ \\
PnPNovelFromTrayToTieredshelf & 15.7 & 32.0 & 31.5 & $26.2\pm3.6$ \\
\midrule
\textbf{Average Success Rate} & 40.4 & 48.0 & 47.6 & $\mathbf{56.5\pm0.4}$ \\
\bottomrule
\end{tabular}
\end{table}

\clearpage

\paragraph{LIBERO.}
Table~\ref{tab:libero_per_task_results} reports the per-task success rates of CoRE-VLA on the four LIBERO suites, including LIBERO-Spatial, LIBERO-Object, LIBERO-Goal, and LIBERO-10/Long. 
Each task is evaluated with 50 rollouts. 
The per-task results show that CoRE-VLA performs consistently across spatial reasoning, object grounding, goal-conditioned manipulation, and long-horizon instruction-following tasks, with only a small performance drop on the more challenging long-horizon suite.

\begin{table*}[!htbp]
\centering
\footnotesize
\setlength{\tabcolsep}{4pt}
\renewcommand{\arraystretch}{0.95}
\caption{Per-task success rates of CoRE-VLA on the four LIBERO suites. Each task is evaluated with 50 rollouts.}
\label{tab:libero_per_task_results}
\begin{tabularx}{\textwidth}{@{}>{\raggedright\arraybackslash}Xc@{}}
\toprule
\textbf{Task} & \textbf{SR (\%)} \\
\midrule

\multicolumn{2}{@{}l}{\textbf{LIBERO-Spatial}} \\
pick up the black bowl between the plate and the ramekin and place it on the plate & 100.0 \\
pick up the black bowl next to the ramekin and place it on the plate & 100.0 \\
pick up the black bowl from table center and place it on the plate & 100.0 \\
pick up the black bowl on the cookie box and place it on the plate & 100.0 \\
pick up the black bowl in the top drawer of the wooden cabinet and place it on the plate & 100.0 \\
pick up the black bowl on the ramekin and place it on the plate & 96.0 \\
pick up the black bowl next to the cookie box and place it on the plate & 100.0 \\
pick up the black bowl on the stove and place it on the plate & 94.0 \\
pick up the black bowl next to the plate and place it on the plate & 100.0 \\
pick up the black bowl on the wooden cabinet and place it on the plate & 100.0 \\
\cmidrule(lr){1-2}
\textbf{Suite Average} & \textbf{99.0} \\

\midrule
\multicolumn{2}{@{}l}{\textbf{LIBERO-Object}} \\
pick up the alphabet soup and place it in the basket & 100.0 \\
pick up the cream cheese and place it in the basket & 100.0 \\
pick up the salad dressing and place it in the basket & 98.0 \\
pick up the bbq sauce and place it in the basket & 100.0 \\
pick up the ketchup and place it in the basket & 100.0 \\
pick up the tomato sauce and place it in the basket & 96.0 \\
pick up the butter and place it in the basket & 100.0 \\
pick up the milk and place it in the basket & 100.0 \\
pick up the chocolate pudding and place it in the basket & 100.0 \\
pick up the orange juice and place it in the basket & 98.0 \\
\cmidrule(lr){1-2}
\textbf{Suite Average} & \textbf{99.2} \\

\midrule
\multicolumn{2}{@{}l}{\textbf{LIBERO-Goal}} \\
open the middle drawer of the cabinet & 100.0 \\
put the bowl on the stove & 100.0 \\
put the wine bottle on top of the cabinet & 96.0 \\
open the top drawer and put the bowl inside & 96.0 \\
put the bowl on top of the cabinet & 98.0 \\
push the plate to the front of the stove & 98.0 \\
put the cream cheese in the bowl & 100.0 \\
turn on the stove & 100.0 \\
put the bowl on the plate & 100.0 \\
put the wine bottle on the rack & 100.0 \\
\cmidrule(lr){1-2}
\textbf{Suite Average} & \textbf{98.8} \\

\midrule
\multicolumn{2}{@{}l}{\textbf{LIBERO-10 / Long}} \\
put both the alphabet soup and the tomato sauce in the basket & 94.0 \\
put both the cream cheese box and the butter in the basket & 100.0 \\
turn on the stove and put the moka pot on it & 98.0 \\
put the black bowl in the bottom drawer of the cabinet and close it & 100.0 \\
put the white mug on the left plate and put the yellow and white mug on the right plate & 100.0 \\
pick up the book and place it in the back compartment of the caddy & 100.0 \\
put the white mug on the plate and put the chocolate pudding to the right of the plate & 90.0 \\
put both the alphabet soup and the cream cheese box in the basket & 100.0 \\
put both moka pots on the stove & 98.0 \\
put the yellow and white mug in the microwave and close it & 96.0 \\
\cmidrule(lr){1-2}
\textbf{Suite Average} & \textbf{97.6} \\

\bottomrule
\end{tabularx}
\end{table*}

\clearpage


\subsection{Details of Real-world Evaluation}
\label{sec:detail_real_world}
\paragraph{Real-world evaluation protocol.}
We report both average task score and average success rate (SR) in Table~\ref{tab:real-world}. 
Each rollout is scored on a 100-point scale according to predefined task-specific subgoals. 
The scoring criteria are as follows:
\begin{itemize}
    \item \textbf{Vegetables-Picking:} The score consists of four 25-point subgoals: grasping the cucumber, placing the cucumber, grasping the eggplant, and placing the eggplant.
    \item \textbf{Clothes-Folding:} The score consists of four 25-point subgoals: folding the left sleeve, folding the right sleeve, and completing two subsequent garment-folding stages.
    \item \textbf{Fabric-Folding:} The score consists of two 50-point subgoals: grasping and lifting the fabric edge, and completing the fold. Fabrics with diverse unseen colors, shapes, and materials are used to evaluate zero-shot generalization.
\end{itemize}
Average success rate measures the percentage of rollouts that complete all task subgoals.
Each method is evaluated over $20$ independent rollouts per task under the same protocol.
\paragraph{Real-world rollout details.}
Tables~\ref{tab:real_robot_vegetables_detailed}, \ref{tab:real_robot_clothes_detailed}, and~\ref{tab:real_robot_fabric_detailed} provide the full rollout-level results for the three real-world tasks: Vegetables-Picking, Clothes-Folding, and Fabric-Folding. 
For each rollout, we report both the task score and the binary success indicator. 
A rollout is considered successful only when it achieves the full end-to-end score of 100. 
These detailed results make the real-world evaluation protocol transparent and show whether failures come from complete task failures or partial completion of intermediate subgoals.
We use Orbbec Dabai DC1 RGB-D cameras as the depth cameras on the AgileX ALOHA, with three units mounted at the left wrist, right wrist, and head.

\begin{table*}[h!]
\centering
\caption{Detailed real-world rollout results on the \textbf{Vegetables-Picking} task. Success indicates whether the rollout achieves an end-to-end score of 100.}
\label{tab:real_robot_vegetables_detailed}
\scriptsize
\setlength{\tabcolsep}{2pt}
\resizebox{\textwidth}{!}{
\begin{tabular}{llcccccccccccccccccccc}
\toprule
\textbf{Model} & \textbf{Metric} 
& \textbf{1} & \textbf{2} & \textbf{3} & \textbf{4} & \textbf{5}
& \textbf{6} & \textbf{7} & \textbf{8} & \textbf{9} & \textbf{10}
& \textbf{11} & \textbf{12} & \textbf{13} & \textbf{14} & \textbf{15}
& \textbf{16} & \textbf{17} & \textbf{18} & \textbf{19} & \textbf{20} \\
\midrule
\multirow{2}{*}{\makecell[l]{$\pi_{0.5}$\\{\scriptsize JAX version}}} & Score
& 100 & 100 & 100 & 100 & 100 & 100 & 0 & 50 & 50 & 100
& 50 & 100 & 75 & 0 & 50 & 75 & 100 & 0 & 100 & 100 \\
& Success
& True & True & True & True & True & True & False & False & False & True
& False & True & False & False & False & False & True & False & True & True \\
\midrule
\multirow{2}{*}{\makecell[l]{Baseline\\{\scriptsize Dense Action DiT w/o depth modality}}} & Score
& 0 & 100 & 100 & 100 & 50 & 50 & 0 & 100 & 50 & 75
& 100 & 0 & 50 & 25 & 50 & 100 & 50 & 100 & 50 & 0 \\
& Success
& False & True & True & True & False & False & False & True & False & False
& True & False & False & False & False & True & False & True & False & False \\
\midrule

\multirow{2}{*}{\makecell[l]{CoRE-VLA\\{\scriptsize Infer. w/o auxiliary depth}}} & Score
& 0 & 100 & 50 & 100 & 100 & 50 & 100 & 50 & 100 & 100
& 100 & 100 & 100 & 100 & 50 & 100 & 100 & 50 & 0 & 100 \\
& Success
& False & True & False & True & True & False & True & False & True & True
& True & True & True & True & False & True & True & False & False & True \\
\midrule
\multirow{2}{*}{\makecell[l]{CoRE-VLA\\{\scriptsize Infer. w/ physical depth camera}}} & Score
& 100 & 50 & 100 & 100 & 100 & 100 & 100 & 50 & 100 & 100
& 100 & 25 & 100 & 0 & 100 & 0 & 50 & 100 & 100 & 100 \\
& Success
& True & False & True & True & True & True & True & False & True & True
& True & False & True & False & True & False & False & True & True & True \\

\bottomrule
\end{tabular}
}
\end{table*}

\begin{table*}[h!]
\centering
\caption{Detailed real-world rollout results on the \textbf{Clothes-Folding} task. Success indicates whether the rollout achieves an end-to-end score of 100.}
\label{tab:real_robot_clothes_detailed}
\scriptsize
\setlength{\tabcolsep}{2pt}
\resizebox{\textwidth}{!}{
\begin{tabular}{llcccccccccccccccccccc}
\toprule
\textbf{Model} & \textbf{Metric} 
& \textbf{1} & \textbf{2} & \textbf{3} & \textbf{4} & \textbf{5}
& \textbf{6} & \textbf{7} & \textbf{8} & \textbf{9} & \textbf{10}
& \textbf{11} & \textbf{12} & \textbf{13} & \textbf{14} & \textbf{15}
& \textbf{16} & \textbf{17} & \textbf{18} & \textbf{19} & \textbf{20} \\
\midrule
\multirow{2}{*}{\makecell[l]{$\pi_{0.5}$\\{\scriptsize JAX version}}} & Score
& 100 & 100 & 100 & 0 & 75 & 0 & 100 & 50 & 100 & 100
& 100 & 100 & 100 & 100 & 0 & 25 & 100 & 75 & 50 & 100 \\
& Success
& True & True & True & False & False & False & True & False & True & True
& True & True & True & True & False & False & True & False & False & True \\
\midrule
\multirow{2}{*}{\makecell[l]{Baseline\\{\scriptsize Dense Action DiT w/o depth modality}}} & Score
& 75 & 75 & 75 & 100 & 75 & 75 & 0 & 75 & 100 & 50
& 100 & 100 & 75 & 50 & 0 & 75 & 100 & 50 & 100 & 0 \\
& Success
& False & False & False & True & False & False & False & False & True & False
& True & True & False & False & False & False & True & False & True & False \\
\midrule
\multirow{2}{*}{\makecell[l]{CoRE-VLA\\{\scriptsize Infer. w/o auxiliary depth}}} & Score
& 100 & 100 & 25 & 100 & 75 & 100 & 50 & 75 & 100 & 75
& 25 & 50 & 50 & 100 & 100 & 100 & 50 & 100 & 75 & 100 \\
& Success
& True & True & False & True & False & True & False & False & True & False
& False & False & False & True & True & True & False & True & False & True \\
\midrule
\multirow{2}{*}{\makecell[l]{CoRE-VLA\\{\scriptsize Infer. w/ physical depth camera}}} & Score
& 100 & 100 & 100 & 100 & 50 & 100 & 75 & 100 & 50 & 0
& 100 & 100 & 75 & 100 & 100 & 0 & 100 & 100 & 25 & 100 \\
& Success
& True & True & True & True & False & True & False & True & False & False
& True & True & False & True & True & False & True & True & False & True \\

\bottomrule
\end{tabular}
}
\end{table*}

\begin{table*}[h!]
\centering
\caption{Detailed real-world rollout results on the \textbf{Fabric-Folding} task. Success indicates whether the rollout achieves an end-to-end score of 100.}
\label{tab:real_robot_fabric_detailed}
\scriptsize
\setlength{\tabcolsep}{2pt}
\resizebox{\textwidth}{!}{
\begin{tabular}{llcccccccccccccccccccc}
\toprule
\textbf{Model} & \textbf{Metric} 
& \textbf{1} & \textbf{2} & \textbf{3} & \textbf{4} & \textbf{5}
& \textbf{6} & \textbf{7} & \textbf{8} & \textbf{9} & \textbf{10}
& \textbf{11} & \textbf{12} & \textbf{13} & \textbf{14} & \textbf{15}
& \textbf{16} & \textbf{17} & \textbf{18} & \textbf{19} & \textbf{20} \\
\midrule
\multirow{2}{*}{\makecell[l]{$\pi_{0.5}$\\{\scriptsize JAX version}}} & Score
& 100 & 100 & 100 & 50 & 100 & 100 & 0 & 0 & 0 & 100
& 100 & 100 & 100 & 0 & 0 & 100 & 100 & 0 & 0 & 100 \\
& Success
& True & True & True & False & True & True & False & False & False & True
& True & True & True & False & False & True & True & False & False & True \\
\midrule
\multirow{2}{*}{\makecell[l]{Baseline\\{\scriptsize Dense Action DiT w/o depth modality}}} & Score
& 50 & 100 & 0 & 100 & 50 & 100 & 100 & 100 & 100 & 0
& 0 & 100 & 50 & 0 & 100 & 50 & 0 & 50 & 100 & 100 \\
& Success
& False & True & False & True & False & True & True & True & True & False
& False & True & False & False & True & False & False & False & True & True \\
\midrule

\multirow{2}{*}{\makecell[l]{CoRE-VLA\\{\scriptsize Infer. w/o auxiliary depth}}} & Score
& 0 & 100 & 100 & 100 & 100 & 100 & 100 & 100 & 100 & 50
& 100 & 100 & 0 & 100 & 100 & 0 & 100 & 100 & 0 & 50 \\
& Success
& False & True & True & True & True & True & True & True & True & False
& True & True & False & True & True & False & True & True & False & False \\
\midrule
\multirow{2}{*}{\makecell[l]{CoRE-VLA\\{\scriptsize Infer. w/ physical depth camera}}} & Score
& 100 & 100 & 100 & 100 & 100 & 100 & 100 & 100 & 100 & 0
& 50 & 100 & 100 & 100 & 100 & 50 & 100 & 50 & 100 & 100 \\
& Success
& True & True & True & True & True & True & True & True & True & False
& False & True & True & True & True & False & True & False & True & True \\

\bottomrule
\end{tabular}
}
\end{table*}

\clearpage

\section{Detailed Ablation Study}
\label{sec:ablation_analysis}
Table~\ref{tab:ablation_analysis} provides a detailed ablation on LIBERO-Long to disentangle the effects of task-intent routing, auxiliary depth perception, modality dropout, modality-specialized experts, and the choice of routing condition. Row 1 is the RGB-language dense baseline with a single general expert and no routing. It achieves an average success rate of $94.4\%$, serving as the reference point for evaluating whether sparse conditional computation and auxiliary-modality extension provide additional benefits.

Row 2 introduces task-intent routing with eight general experts, while still using only RGB-language inputs. Compared with Row 1, the success rate improves from $94.4\%$ to $95.2\%$. Since no depth perception, modality dropout, or modality-specialized experts are used in either row, this comparison isolates the effect of routing among multiple general experts. The improvement suggests that even without auxiliary perception, task-intent-conditioned routing can allocate action-generation computation more effectively across different long-horizon subgoals, supporting our claim that CoRE improves task scalability through specialized yet reusable computation paths.

Rows 3--4 evaluate a naive depth-augmented variant with task-intent routing and general experts, but without modality dropout or modality-specialized experts. Row 3 enables depth at inference and achieves $95.6\%$, slightly higher than the RGB-only routed model in Row 2. This indicates that auxiliary depth perception can provide useful geometric cues for long-horizon manipulation. However, Row 4 removes depth at inference while using the same model trained with depth, causing the success rate to drop sharply to $91.2\%$. The comparison between Rows 3 and 4 shows that simply adding depth perception can make the policy entangle depth-dependent features with the general action-generation pathway, leading to poor robustness when the auxiliary modality is unavailable. This result motivates the need for explicit missing-modality training and modality-aware routing.

Rows 5--6 add modality dropout to the depth-augmented routed model, but still use only general experts without modality-specialized experts. Compared with Row 3, Row 5 obtains a lower depth-enabled success rate of $94.4\%$, suggesting that modality dropout alone may regularize the model toward a more conservative RGB-language-compatible representation and reduce the immediate benefit obtained from depth when there is no dedicated modality-specific capacity. However, Row 6 achieves $96.2\%$ when depth is disabled at inference, substantially improving over Row 4 ($91.2\%$). This comparison shows that modality dropout is effective for preserving a non-depth inference path. At the same time, the contrast between Row 5 and Row 6 indicates that dropout alone does not guarantee effective exploitation of depth cues; without modality-specialized experts, the model lacks a separate computation pathway to absorb auxiliary-modality-specific information.

Rows 7--8 introduce modality-specialized experts but do not use modality dropout. Row 7 enables depth at inference and achieves $94.6\%$, while Row 8 disables depth at inference and drops to $92.8\%$. Compared with Rows 3--4, adding modality-specialized experts without modality dropout slightly improves missing-depth robustness from $91.2\%$ to $92.8\%$, but the performance remains clearly below the full model. This suggests that modality-specialized experts alone are not sufficient: although they provide additional capacity for depth-dependent computation, the policy is not explicitly trained to operate under both full-modality and RGB-only regimes. As a result, the action generator can still over-rely on depth-related computation and remains fragile when depth is removed.

Rows 9--10 further examine the role of the routing condition. Row 9 keeps depth perception, modality dropout, and modality-specialized experts, but removes task-intent routing. It achieves $95.4\%$, which is better than Row 7 ($94.6\%$) because modality dropout improves robustness and regularization, but still lower than the full CoRE-VLA model in Row 11 ($97.6\%$). This comparison shows that the gains of CoRE-VLA do not come merely from adding more experts, depth features, or modality dropout. Task-intent routing is necessary for assigning action-side computation to experts according to the current language instruction and manipulation context.

Row 10 keeps the same modality-aware design as the full model but changes the routing condition from the text-intent embedding to a pooled full vision-language context, i.e., $g=\operatorname{AvgPool}(C_{\mathrm{vl}})$. This variant achieves only $94.2\%$, which is $3.4$ points lower than the full model in Row 11. This result highlights that the source of the routing signal is crucial. In VLA representations, visual tokens typically substantially outnumber text tokens, so pooling the full multimodal sequence can make the resulting routing signal dominated by scene-level visual information, including object appearance, background, and geometry, thereby diluting the instruction-level task intent that is most relevant for action specialization. In contrast, CoRE-VLA uses $g=\operatorname{AvgPool}(C_{\mathrm{text}})$, where the VLM-encoded text tokens preserve an instruction-centered task representation while remaining grounded in the current visual context through multimodal attention. The comparison between Rows 10 and 11 therefore supports our design choice of using text-token pooling as an explicit task-intent signal to guide expert routing in the action head, rather than directly pooling the visual-token-dominated multimodal sequence.

Rows 11--12 correspond to the full CoRE-VLA configuration, combining task-intent routing, modality dropout, and modality-specialized experts. Row 11 achieves the best success rate of $97.6\%$ when depth is available at inference, showing that the full model can effectively exploit auxiliary geometric information. Row 12 disables depth at inference and still maintains a high success rate of $97.0\%$, only $0.6$ points below Row 11. This small gap contrasts sharply with the naive depth-augmented setting in Rows 3--4, where removing depth causes a $4.4$-point drop. Therefore, the full CoRE-VLA design successfully benefits from auxiliary depth perception while preserving a reliable non-depth computation path.

Overall, the ablation supports four central claims of CoRE-VLA. First, Rows 1--2 show that task-intent routing improves long-horizon action generation even without auxiliary modalities. Second, Rows 3--8 show that auxiliary depth perception, modality dropout, and modality-specialized experts address different aspects of auxiliary-depth utilization and robustness: depth provides useful geometric cues, dropout trains the model to handle missing modalities, and modality-specialized experts provide capacity for separating auxiliary-modality-dependent computation from general manipulation computation. Third, Rows 9--11 show that the full gain requires task-intent-conditioned routing rather than merely adding experts or auxiliary modalities. Fourth, Row 10 demonstrates that not all routing conditions are equally effective: using a pooled full vision-language representation as the routing signal substantially underperforms the text-intent-conditioned router, validating our choice of instruction-centered routing for long-horizon manipulation.

\begin{table}[h!]
\centering
\footnotesize
\caption{Detailed ablation analysis on the LIBERO-Long benchmark.}
\label{tab:ablation_analysis}
\begin{tabular}{c c c c cc c c}
\toprule
\textbf{No.}
& \makecell[c]{Depth\\Perception}
& \makecell[c]{Task-intent\\Routing}
& \makecell[c]{Modality\\Dropout}
& $|\mathcal{E}_{\mathrm{gen}}|$
& $|\mathcal{E}_{\mathrm{mod}}|$
& \makecell[c]{Enable Depth\\ at Inference}
& \makecell[c]{\textbf{Avg. SR}} \\
\midrule
1 & --         & --         & --         & 1 & 0 & --         & 94.4   \\
2 & --         & \checkmark & --         & 8 & 0 & --         & 95.2   \\

\addlinespace
3 & \checkmark & \checkmark & --         & 8 & 0 & \checkmark & 95.6   \\
4 & \checkmark & \checkmark & --         & 8 & 0 & --         & 91.2   \\

\addlinespace
5 & \checkmark & \checkmark & \checkmark & 8 & 0 & \checkmark & 94.4   \\
6 & \checkmark & \checkmark & \checkmark & 8 & 0 & --         & 96.2   \\

\addlinespace
7 & \checkmark & \checkmark & --         & 8 & 4 & \checkmark & 94.6   \\
8 & \checkmark & \checkmark & --         & 8 & 4 & --         & 92.8   \\

\addlinespace
9 & \checkmark & --         & \checkmark & 8 & 4 & \checkmark & 95.4   \\
10 & \checkmark & $g=\operatorname{AvgPool}(C_{\mathrm{vl}})$  & \checkmark & 8 & 4 & \checkmark & 94.2   \\
11 & \checkmark & \checkmark & \checkmark & 8 & 4 & \checkmark & \textbf{97.6} \\
12 & \checkmark & \checkmark & \checkmark & 8 & 4 & --         & 97.0   \\
\bottomrule
\end{tabular}
\end{table}

\clearpage

\section{Router Analysis}
\label{sec:router_analysis}

\paragraph{Expert utilization.}
Expert utilization is computed over routed token-dispatches in CoRE blocks.
For a given evaluation subset, such as a task or suite, we aggregate all selected action-side DiT tokens across CoRE layers, diffusion denoising steps, rollout timesteps, and episodes.
The utilization of expert $e$ is defined as
\begin{equation}
    U_e = \frac{N_e}{\sum_j N_j},
\end{equation}
where $N_e$ denotes the number of selected token-dispatches routed to expert $e$, and the denominator is the total number of selected token-dispatches routed to all experts in the same subset.
Thus, the reported expert usage reflects aggregate routing and compute allocation rather than the utilization of a single CoRE layer.

To better understand the behavior of the learned router, we analyze CoRE-VLA from three complementary perspectives.
First, we visualize how expert usage evolves over rollout time to examine whether routing changes with task progress.
Second, we align expert utilization and token-selection activity with rollout frames to inspect whether different execution phases activate different computation paths.
Third, we compare task-level expert-utilization patterns to study whether the router learns task-dependent expert allocation rather than using a fixed or collapsed routing distribution.
Together, these analyses provide qualitative evidence that CoRE routing induces structured, dynamic, and task-conditioned sparse computation inside the action generator.

\begin{figure}[h!]
    \centering
    \includegraphics[width=\linewidth]{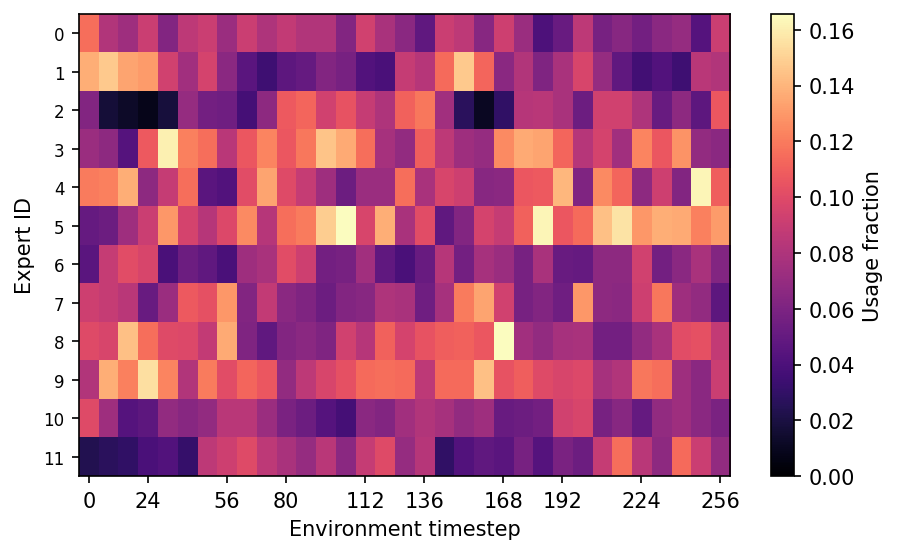}
    \vspace{-10pt}
    \caption{Temporal evolution of expert utilization during policy rollout on LIBERO-Long.
Each column shows the normalized expert-usage distribution at an environment timestep, aggregated over CoRE layers and diffusion denoising steps.
The routing pattern changes over time, suggesting that CoRE-VLA dynamically reallocates action-side computation as the task progresses through different execution stages and subgoals.}
    \label{fig:routing_timeline_libero10}
\end{figure}

\begin{figure}[h!]
    \centering
    \includegraphics[width=\linewidth]{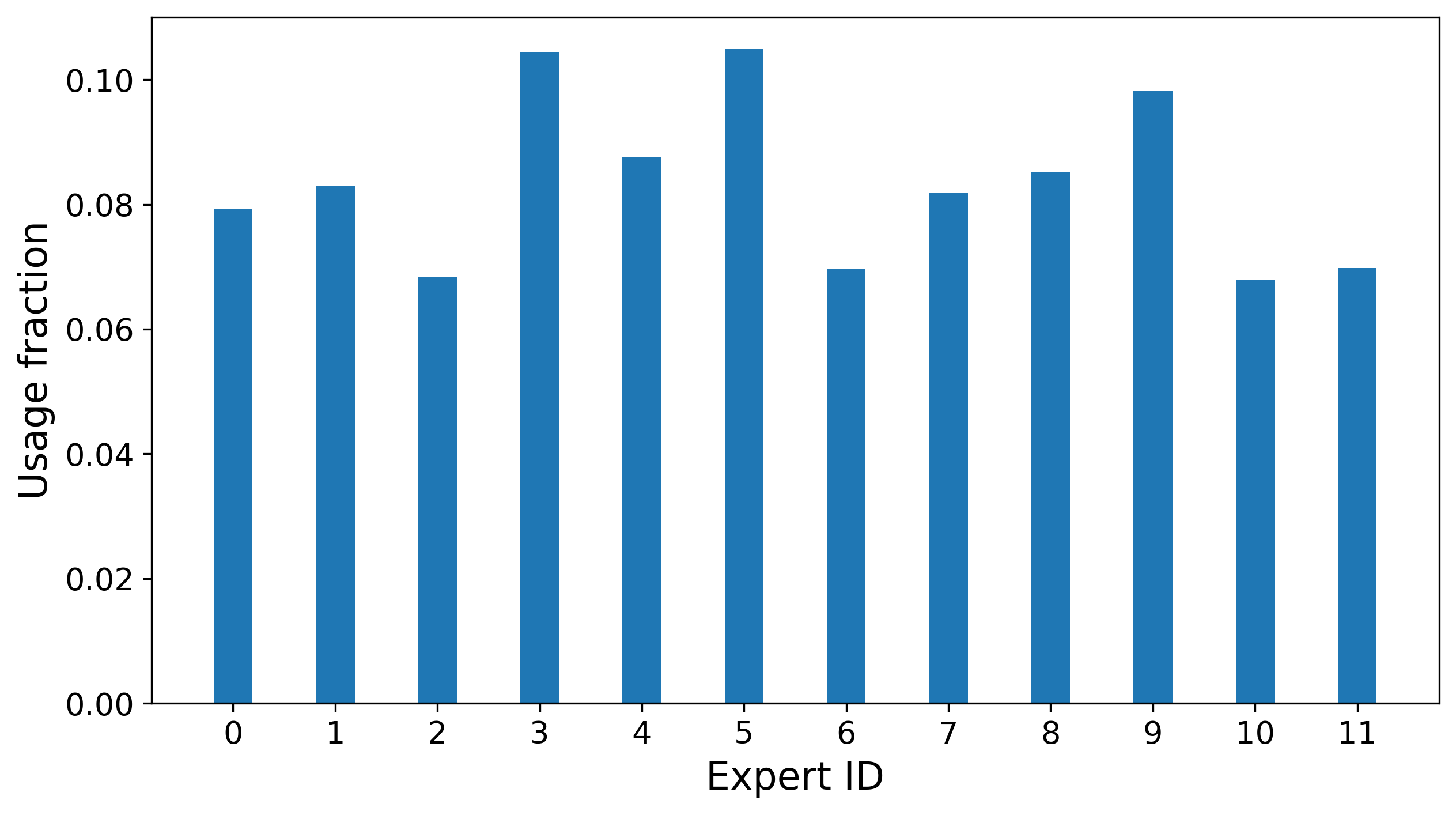}
    \vspace{-10pt}
    \caption{Overall expert utilization across LIBERO-Long tasks.
Expert usage is computed as the normalized fraction of routed action-side token-dispatches assigned to each expert, aggregated over CoRE layers, diffusion denoising steps, rollout timesteps, and episodes.
The distribution is moderately balanced but non-uniform, indicating that CoRE-VLA avoids severe expert collapse while selectively allocating more computation to frequently useful experts.}
    \label{fig:expert_usage_bar_libero10}
\end{figure}

\paragraph{Temporal routing dynamics on LIBERO-Long.}
We first analyze routing behavior on LIBERO-Long, which contains multi-stage manipulations and therefore provides a suitable setting for studying whether the router changes its computation pattern as the task progresses, as shown in Figures~\ref{fig:routing_timeline_libero10} and~\ref{fig:expert_usage_bar_libero10}.
The temporal expert-utilization map shows the normalized expert distribution at each environment timestep, while the overall usage plot summarizes the aggregate allocation across the full rollout.
These two views are complementary: the timeline reveals when routing changes, whereas the aggregate distribution shows whether the learned router avoids degenerate expert collapse.

\begin{figure}[h!]
    \centering
    \includegraphics[width=1\linewidth]{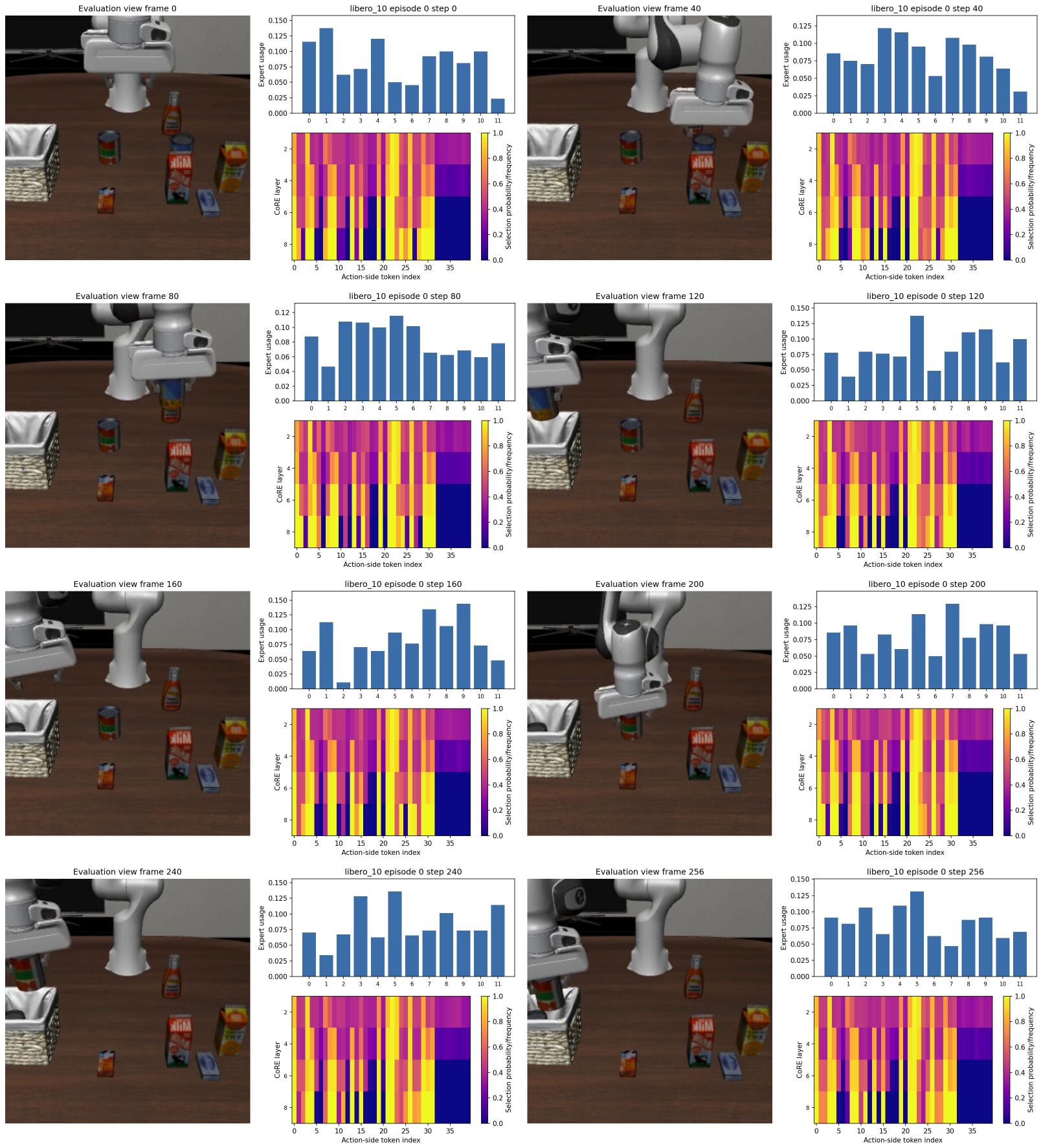}
    \caption{Long-horizon CoRE routing dynamics with video-aligned visualization on LIBERO-Long. Each row shows two rollout timesteps aligned with the corresponding evaluation video frames. The upper routing panel shows the normalized expert utilization at that timestep, aggregated over selected action-side token dispatches across CoRE layers and diffusion denoising steps. The lower heatmap shows CoRE token-selection activity over the action-side DiT sequence, where the y-axis denotes CoRE layer index and the x-axis denotes action-side token representation id. In the LIBERO setting, token id 0 corresponds to the proprioceptive state token, token ids 1--32 correspond to learnable action-query tokens, and token ids 33--40 correspond to noisy action-trajectory tokens for future action steps 0--7. Expert ids 0-7 correspond to general experts, and Expert ids 8-11 correspond to modality-specialized experts. The visualization shows that different phases of the long-horizon rollout activate different expert mixtures and different subsets of action-side representations, suggesting task-progress-dependent conditional computation rather than a static routing pattern.}
    \label{fig:libero10_episode_0_grid}
\end{figure}

\paragraph{Video-aligned routing and token selection.}
To connect routing decisions with the actual execution process, we further visualize CoRE routing together with video frames from the same rollout, as shown in Figure~\ref{fig:libero10_episode_0_grid}.
For each selected timestep, we show both the expert-utilization distribution and the token-selection heatmap over the action-side DiT sequence.
This visualization helps distinguish two levels of conditional computation: which experts are used, and which action-side token representations are selected for routed computation.
The changing expert mixtures and token-selection patterns across rollout stages indicate that CoRE-VLA does not use a static sparse path, but instead adapts computation according to task progress and the currently required manipulation behavior.

\begin{figure}[h!]
    \centering
    \includegraphics[width=0.85\linewidth]{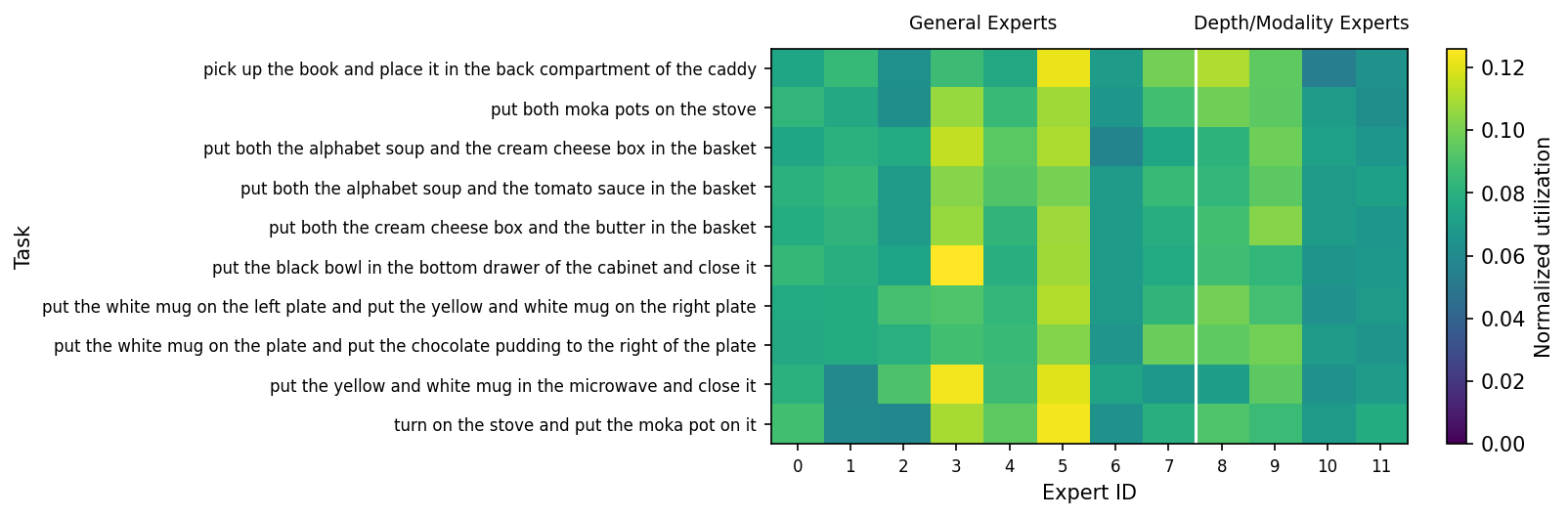}
    \vspace{-12pt}
    \caption{Expert utilization heatmap of diverse tasks on LIBERO-Long.}
    \label{fig:task_expert_utilization_heatmap_libero10}
\end{figure}

\begin{figure}[h!]
    \centering
    \includegraphics[width=0.85\linewidth]{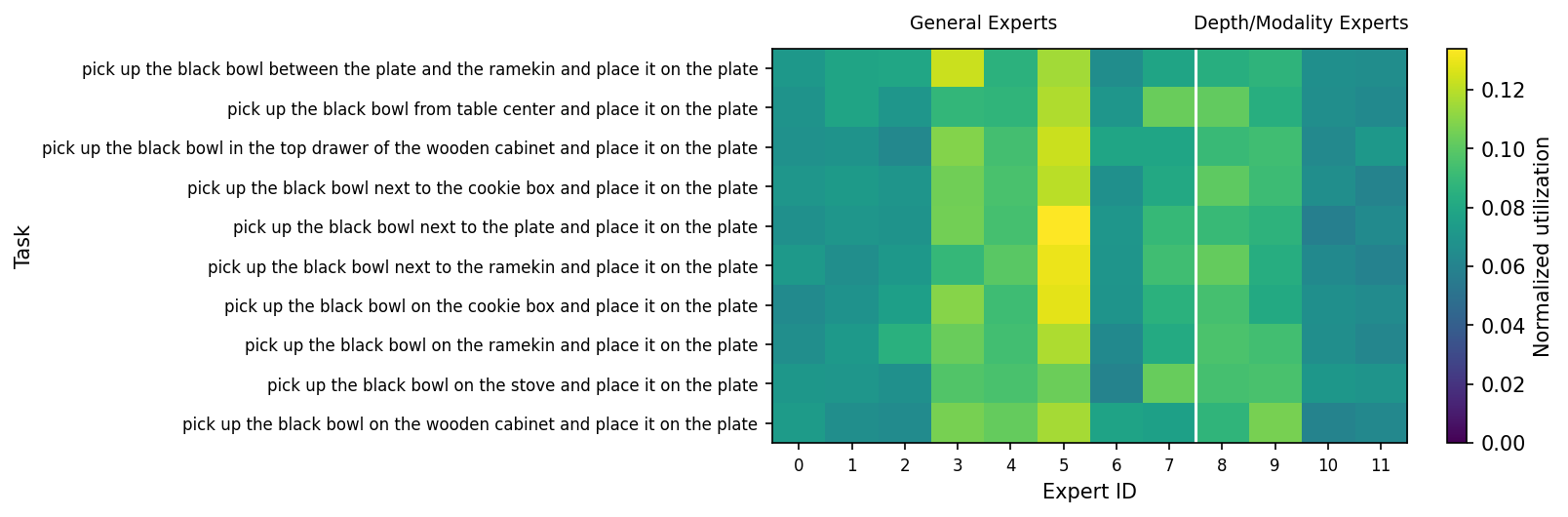}
    \vspace{-12pt}
    \caption{Expert utilization heatmap of diverse tasks on LIBERO-Spatial.}
    \label{fig:task_expert_utilization_heatmap_libero_spatial}
\end{figure}

\begin{figure}[h!]
    \centering
    \includegraphics[width=0.8\linewidth]{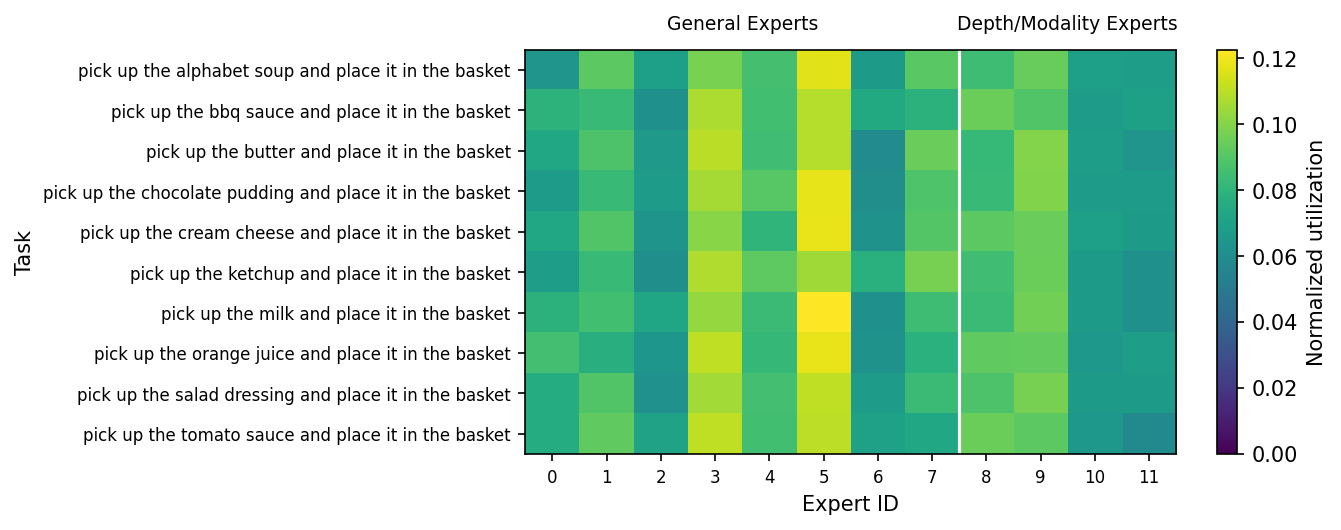}
    \vspace{-12pt}
    \caption{Expert utilization heatmap of diverse tasks on LIBERO-Object.}
    \label{fig:task_expert_utilization_heatmap_libero_object}
\end{figure}

\begin{figure}[h!]
    \centering
    \includegraphics[width=0.8\linewidth]{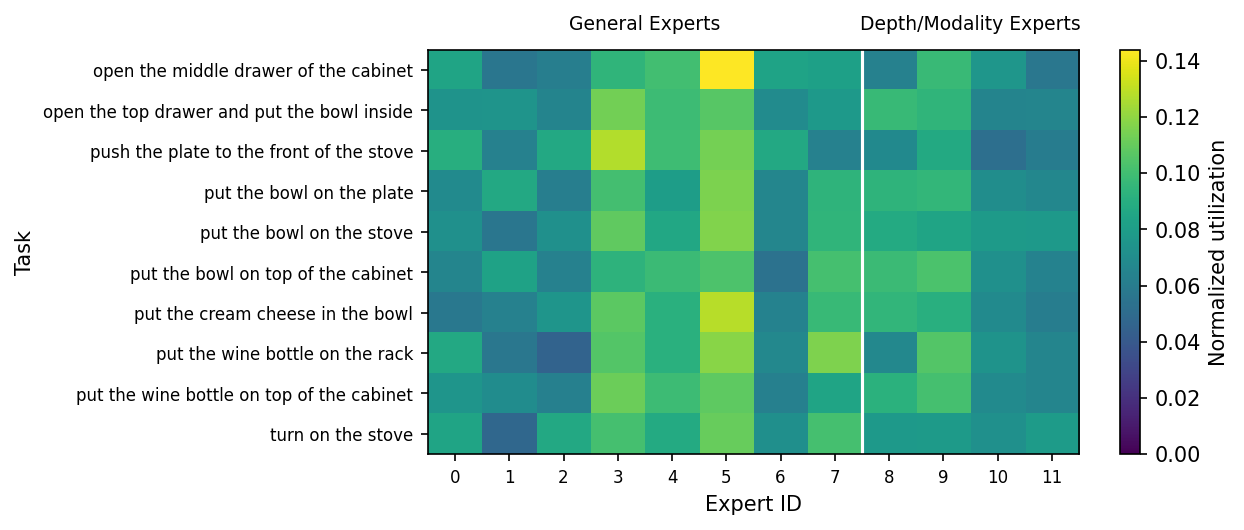}
    \vspace{-12pt}
    \caption{Expert utilization heatmap of diverse tasks on LIBERO-Goal.}
    \label{fig:task_expert_utilization_heatmap_libero_goal}
\end{figure}

\paragraph{Task-dependent expert utilization across LIBERO suites.}
We next aggregate expert utilization at the task level and visualize the resulting task-by-expert usage matrix for each LIBERO suite as shown in Figure~\ref{fig:task_expert_utilization_heatmap_libero10}--\ref{fig:task_expert_utilization_heatmap_libero_goal}.
Each row corresponds to a task instruction, and each column corresponds to an expert.
If the router ignored task intent or collapsed to a fixed allocation, different tasks would exhibit nearly identical expert-usage patterns.
Instead, the heatmaps show structured similarities as well as task-specific differences across rows, suggesting that the router learns both reusable action-processing experts and task-adaptive expert allocation.

Overall, the task-level heatmaps show structured yet task-dependent routing patterns.
Some experts are repeatedly activated across different tasks, suggesting reusable action-processing patterns shared by multiple manipulation behaviors.
At the same time, expert-utilization distributions vary across task instructions and suites, indicating that CoRE-VLA adapts its computation according to task intent rather than relying on a fixed expert assignment.
These results support the interpretation that routed experts capture both general manipulation primitives and task-specialized action-processing modes.

\paragraph{Routing dynamics on RoboCasa GR1 Tabletop.}
We also visualize routing behavior on RoboCasa GR1 Tabletop, using the task \textit{PnPBottleToCabinetClose} as a representative long-horizon pick-and-place example.

\begin{figure}[h!]
    \centering
    \includegraphics[width=0.9\linewidth]{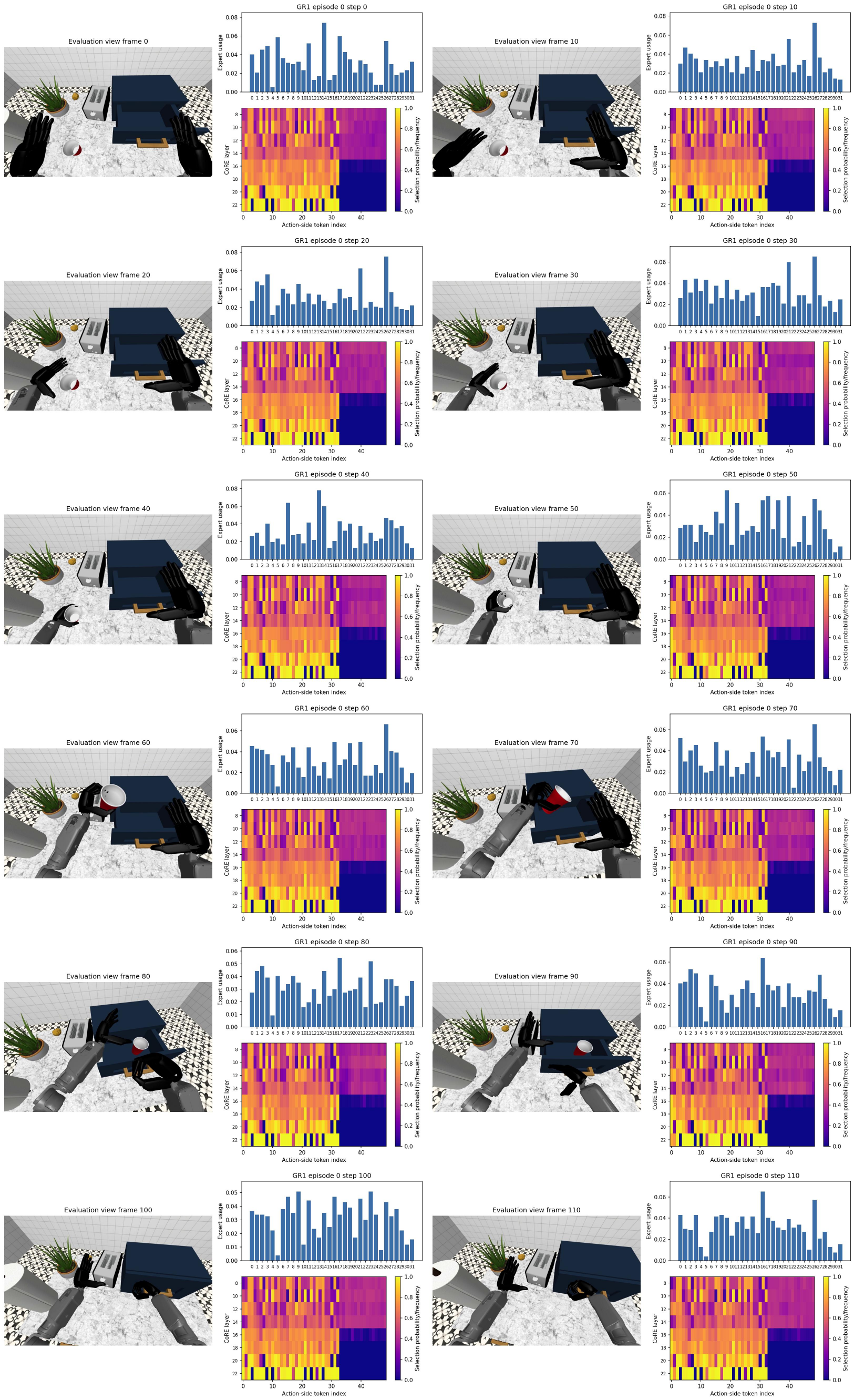}
    \caption{Long-horizon CoRE routing dynamics with video-aligned visualization on RoboCasa GR1 Tabletop task \textit{PnPBottleToCabinetClose}. Token id 0 corresponds to the proprioceptive state token, token ids 1--32 correspond to learnable action-query tokens, and token ids 33--48 correspond to the noisy action-trajectory tokens for future action steps 0--15. Expert ids 0--23 correspond to general experts, and Expert ids 24--31 correspond to modality-specialized experts.}
    \label{fig:GR1_episode_0_grid}
\end{figure}

Compared with LIBERO, RoboCasa GR1 uses a larger action generator and more experts, making it useful for inspecting whether CoRE routing remains structured under a higher-capacity policy.
As shown in Figure~\ref{fig:GR1_episode_0_grid}, the video-aligned visualization shows how expert usage and token-selection activity evolve as the policy approaches, grasps, transports, and places the object.

\begin{figure}
    \centering
    \includegraphics[width=1\linewidth]{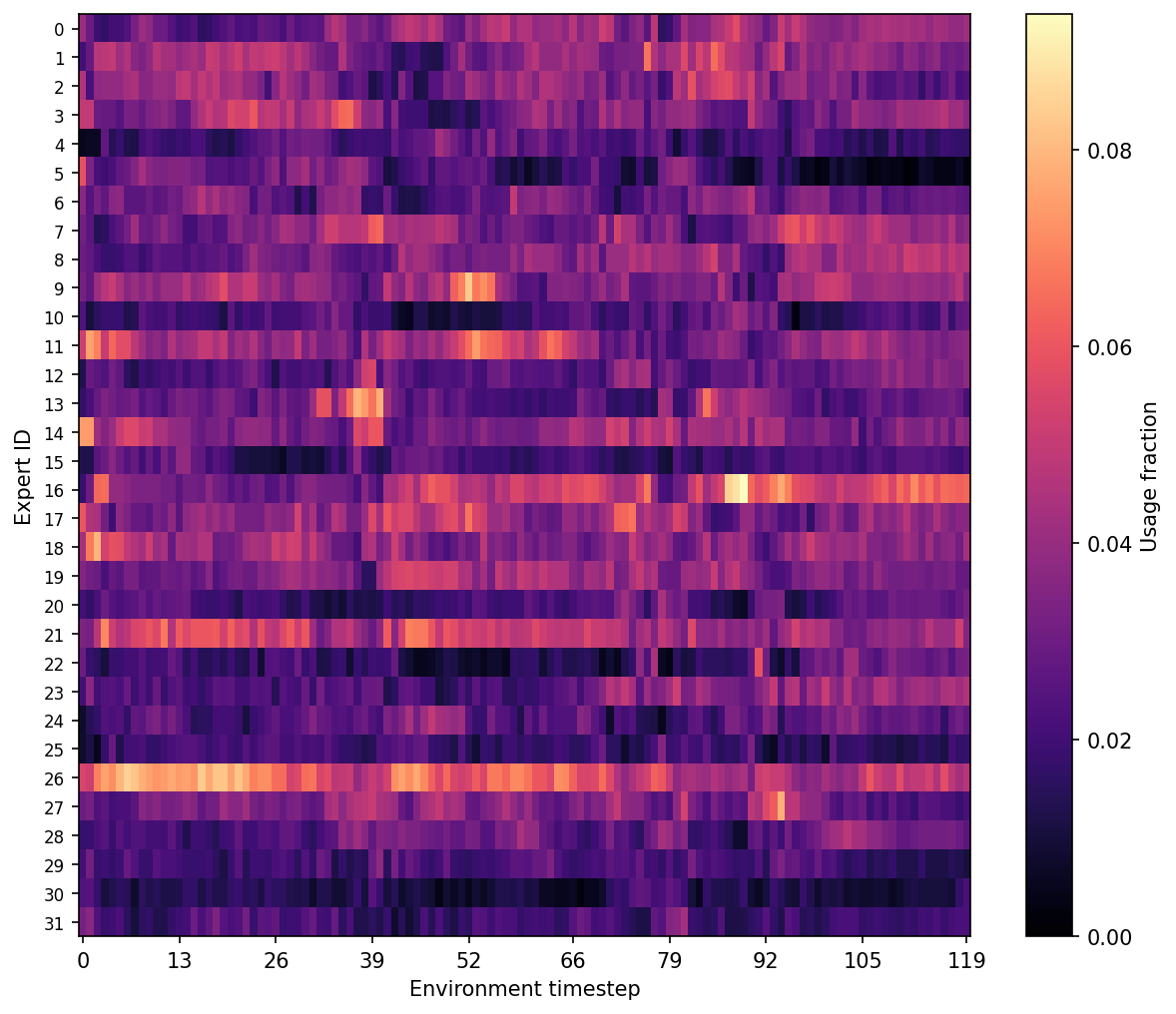}
    \vspace{-10pt}
    \caption{Temporal evolution of expert utilization during policy rollout on RoboCasa GR1 Tabletop task \textit{PnPBottleToCabinetClose}.
Each column shows the normalized expert-usage distribution at an environment timestep, aggregated over CoRE layers and diffusion denoising steps.
The routing pattern changes over time, suggesting that CoRE-VLA dynamically reallocates action-side computation as the task progresses through different execution stages and subgoals.}
    \label{fig:long_horizon_routing_timeline_gr1}
\end{figure}

\begin{figure}
    \centering
    \includegraphics[width=1\linewidth]{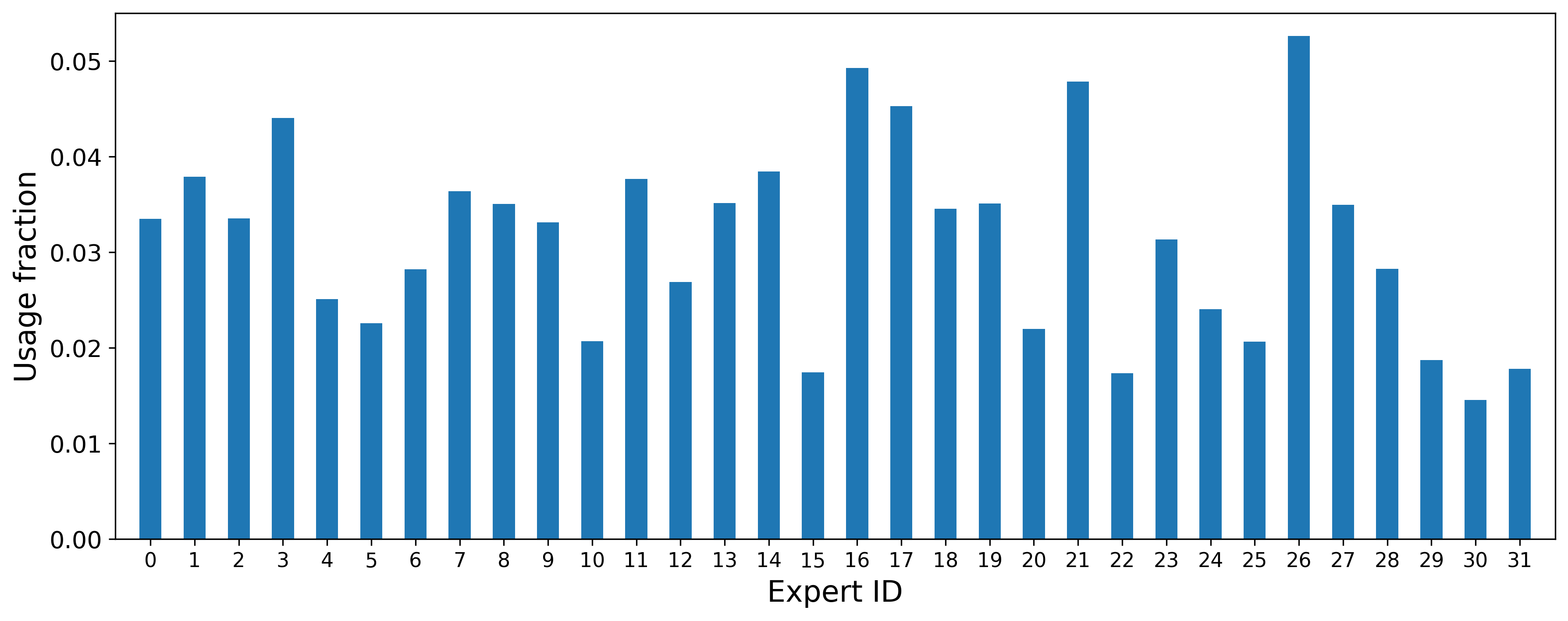}
    \vspace{-10pt}
    \caption{Overall expert utilization on RoboCasa GR1 Tabletop task \textit{PnPBottleToCabinetClose}.
Expert usage is computed as the normalized fraction of routed action-side token-dispatches assigned to each expert, aggregated over CoRE layers, diffusion denoising steps, rollout timesteps, and episodes.
The distribution is moderately balanced but non-uniform, indicating that CoRE-VLA avoids severe expert collapse while selectively allocating more computation to frequently useful experts.}
    \label{fig:rgbd_expert_usage_gr1}
\end{figure}

As shown in Figure~\ref{fig:long_horizon_routing_timeline_gr1}, the temporal routing map further summarizes how expert allocation changes over the full RoboCasa rollout.
As shown in Figure~\ref{fig:rgbd_expert_usage_gr1}, the non-uniform but distributed expert usage indicates that the router selectively emphasizes different experts without collapsing to a small fixed subset.
Together with the video-aligned visualization, these results suggest that CoRE-VLA performs task-progress-dependent computation allocation in both LIBERO and RoboCasa settings.

In summary, the routing visualizations provide three observations.
First, expert utilization changes over rollout time, indicating dynamic computation allocation during long-horizon execution.
Second, token-selection activity varies across CoRE layers and action-side token types, showing that CoRE sparsifies not only expert computation but also the action representations that receive routed processing.
Third, task-level heatmaps exhibit both shared and task-specific expert usage, suggesting that the learned experts support a mixture of reusable manipulation primitives and task-conditioned specialization.
These observations are consistent with the design goal of CoRE-VLA: scaling action generation through conditional computation rather than uniformly applying the same dense computation path to all tasks and rollout stages.

\clearpage

\clearpage

\clearpage

\clearpage

\clearpage

\clearpage

\section{Efficiency Analysis}
\label{app:efficiency}

CoRE-VLA is designed to increase the conditional computation capacity of the action generator without activating all added parameters for every input. In this section, we analyze this capacity--compute trade-off from the perspective of action-side token selection, expert routing, and end-to-end action-generator FLOPs.

\paragraph{Notation.}
Let $L_a$ denote the number of action-side tokens in the Action DiT, $L_c$ the number of conditioning tokens, $d$ the hidden dimension, and $\mathcal{I}_{\mathrm{CoRE}}$ the set of transformer layers equipped with CoRE blocks. We denote $N_{\mathrm{CoRE}} = |\mathcal{I}_{\mathrm{CoRE}}|$. At CoRE layer $\ell$, the selector activates
\[
k_\ell = \max(1, \lfloor \rho L_a \rfloor)
\]
tokens, where $\rho \in (0,1]$ is the selection-capacity ratio. Let
\[
\mathcal{E}_{\mathrm{valid}}(z_{\mathrm{mod}})
=
\{e \mid M_e(z_{\mathrm{mod}})=1\}
\]
be the set of experts available under modality indicator $z_{\mathrm{mod}}$. When the auxiliary modality is unavailable, modality-specialized experts are masked and only general experts remain valid.

\paragraph{Path capacity induced by token selection and expert routing.}
A dense action generator applies the same computation path to every token and every task. In contrast, a CoRE layer first chooses a subset of action-side tokens and then assigns each selected token to one valid expert. Therefore, at layer $\ell$, the number of possible sequence-level routed computation patterns is
\[
|\mathcal{P}_{\ell}(z_{\mathrm{mod}})|
=
\binom{L_a}{k_\ell}
|\mathcal{E}_{\mathrm{valid}}(z_{\mathrm{mod}})|^{k_\ell}.
\]
Across $N_{\mathrm{CoRE}}$ CoRE layers, the number of possible routed computation patterns becomes
\[
|\mathcal{P}_{\mathrm{seq}}(z_{\mathrm{mod}})|
=
\prod_{\ell \in \mathcal{I}_{\mathrm{CoRE}}}
\binom{L_a}{k_\ell}
|\mathcal{E}_{\mathrm{valid}}(z_{\mathrm{mod}})|^{k_\ell}.
\]
Equivalently, from the perspective of a single token, each CoRE layer provides either a bypass path or one of the valid expert paths, yielding
\[
|\mathcal{P}_{\mathrm{token}}(z_{\mathrm{mod}})|
=
\left(
|\mathcal{E}_{\mathrm{valid}}(z_{\mathrm{mod}})| + 1
\right)^{N_{\mathrm{CoRE}}}.
\]
Thus, increasing the number of experts enlarges the conditional action-processing path space multiplicatively, while modality masking changes the path space according to the available sensor set.

\paragraph{Activated computation of a CoRE block.}
Let $C_{\mathrm{cond}}(L_c,d)$ denote the per-token cost of the conditional computation applied to a selected action-side token, including cross-attention to the conditioning sequence and one routed expert transformation. A dense conditional block applies this computation to all $L_a$ action-side tokens:
\[
F_{\mathrm{dense}}^{\ell}
=
L_a \, C_{\mathrm{cond}}(L_c,d).
\]
By contrast, CoRE applies the expensive conditional computation only to the selected $k_\ell$ tokens. The remaining tokens bypass cross-attention and expert computation and are merged back through scatter--merge. The FLOPs of a CoRE layer can therefore be written as
\[
F_{\mathrm{CoRE}}^{\ell}
=
k_\ell \, C_{\mathrm{cond}}(L_c,d)
+
F_{\mathrm{route}}^{\ell},
\]
where $F_{\mathrm{route}}^{\ell}$ contains the lightweight selector, expert-routing logits, masking, and scatter--merge overhead. Since $k_\ell \approx \rho L_a$, we have
\[
\frac{
F_{\mathrm{CoRE}}^{\ell}
}{
F_{\mathrm{dense}}^{\ell}
}
=
\rho
+
\epsilon_\ell,
\quad
\epsilon_\ell
=
\frac{
F_{\mathrm{route}}^{\ell}
}{
L_a \, C_{\mathrm{cond}}(L_c,d)
}.
\]
The overhead term $\epsilon_\ell$ is small when the conditional computation is dominated by cross-attention and FFN/expert transformations. This shows that CoRE bounds the activated conditional computation by the selection-capacity ratio $\rho$, up to routing overhead.

\paragraph{Decoupling parameter capacity from activated compute.}
Suppose each expert has parameter size $P_{\mathrm{exp}}$ and per-token computation cost $C_{\mathrm{exp}}$. Adding more experts increases the total parameter capacity approximately linearly:
\[
P_{\mathrm{experts}}
=
|\mathcal{E}| P_{\mathrm{exp}}.
\]
However, under top-1 token-choice routing, each selected token activates only one expert. Therefore, the expert computation at layer $\ell$ scales as
\[
F_{\mathrm{expert}}^{\ell}
=
k_\ell C_{\mathrm{exp}}
\approx
\rho L_a C_{\mathrm{exp}},
\]
rather than $|\mathcal{E}| L_a C_{\mathrm{exp}}$. The number of experts mainly affects the routing space and total stored parameters, while the activated expert computation remains proportional to the number of selected token-dispatches. This decouples model capacity from per-input activated computation.

\paragraph{End-to-end action-generator FLOPs.}
Let $F_{\mathrm{other}}$ denote the computation of all parts of the Action DiT that are unchanged by CoRE, including non-CoRE layers and shared projections. The dense action generator has cost
\[
F_{\mathrm{dense,total}}
=
F_{\mathrm{other}}
+
\sum_{\ell \in \mathcal{I}_{\mathrm{CoRE}}}
F_{\mathrm{dense}}^{\ell}.
\]
The CoRE action generator has cost
\[
F_{\mathrm{CoRE,total}}
=
F_{\mathrm{other}}
+
\sum_{\ell \in \mathcal{I}_{\mathrm{CoRE}}}
F_{\mathrm{CoRE}}^{\ell}.
\]
Define
\[
\eta
=
\frac{
\sum_{\ell \in \mathcal{I}_{\mathrm{CoRE}}}
F_{\mathrm{dense}}^{\ell}
}{
F_{\mathrm{dense,total}}
}
\]
as the fraction of dense action-generator FLOPs located in the layers and operations replaced by CoRE. Substituting the block-level bound gives
\[
\frac{
F_{\mathrm{CoRE,total}}
}{
F_{\mathrm{dense,total}}
}
\le
1 - (1-\rho)\eta + \bar{\epsilon},
\]
where $\bar{\epsilon}$ aggregates routing and scatter--merge overhead across CoRE layers. This bound clarifies two points. First, the ideal saving is controlled by the capacity ratio $\rho$: smaller $\rho$ activates fewer token-level conditional computations. Second, the realized end-to-end speedup also depends on $\eta$, because only the CoRE-equipped part of the action generator is sparsified.

The above analysis focuses on the action generator, where CoRE is inserted. It does not claim to reduce the computation of the pretrained vision-language backbone or the auxiliary modality encoder. In full closed-loop deployment, end-to-end latency is therefore affected by the VLM, auxiliary depth estimation, diffusion steps, and system overhead.

\clearpage

\section{Formal Derivation of CoRE-VLA with Flow Matching}
\label{app:formal_core_flow_matching}

This section provides a formal derivation of CoRE-VLA as a conditional velocity-field model for action-chunk generation. We first formulate flow matching for language-conditioned robotic actions, and then show how Conditional Routing of Experts (CoRE) parameterizes the velocity field through task- and modality-adaptive sparse computation.

\subsection{Problem Setup}

At control step $t$, the policy receives RGB observations $I_t$, a language instruction $\ell_{\mathrm{lang}}$, and a proprioceptive state $s_t$. When an auxiliary perceptual modality is available, such as estimated depth, tactile sensing, or force feedback, we denote it by $A_t$. The goal is to predict a future action chunk
\begin{equation}
    \mathbf{a}_{t:t+H}
    =
    \{a_t, a_{t+1}, \ldots, a_{t+H}\}
    \in \mathbb{R}^{(H+1)\times d_a},
\end{equation}
where $H$ is the action horizon and $d_a$ is the action dimension.

The vision-language backbone encodes the RGB observation and language instruction into multimodal tokens:
\begin{equation}
    Z_{\mathrm{vl}} = \Phi_{\mathrm{VLM}}(I_t, \ell_{\mathrm{lang}}).
\end{equation}
Let $C_{\mathrm{text}} \subset Z_{\mathrm{vl}}$ denote the subset of VLM-encoded tokens corresponding to the language instruction. We extract a task-intent embedding by average pooling:
\begin{equation}
    g = \operatorname{AvgPool}(C_{\mathrm{text}}) \in \mathbb{R}^{d_c}.
\end{equation}
Because these text tokens are processed by the vision-language backbone, $g$ represents the language-specified task intent grounded in the visual context.

We use a binary modality indicator $z_{\mathrm{mod}}\in\{0,1\}$ to indicate whether the auxiliary modality is enabled. During training, modality dropout samples
\begin{equation}
    z_{\mathrm{mod}} \sim \operatorname{Bernoulli}(1-p_{\mathrm{drop}}),
\end{equation}
where $p_{\mathrm{drop}}$ is the auxiliary-modality dropout probability. If $z_{\mathrm{mod}}=1$, the auxiliary modality encoder produces
\begin{equation}
    Z_{\mathrm{mod}} = \Phi_{\mathrm{mod}}(A_t),
\end{equation}
and the condition sequence is
\begin{equation}
    C(z_{\mathrm{mod}}) = [Z_{\mathrm{mod}}; Z_{\mathrm{vl}}].
\end{equation}
If $z_{\mathrm{mod}}=0$, auxiliary modality tokens are removed and the condition sequence becomes
\begin{equation}
    C(z_{\mathrm{mod}}) = Z_{\mathrm{vl}}.
\end{equation}
At inference time, $z_{\mathrm{mod}}$ is set according to whether the deployed robot has access to the auxiliary sensor.

\subsection{Conditional Flow Matching for Action-Chunk Generation}

CoRE-VLA models action generation as learning a conditional velocity field that transports Gaussian noise to the data distribution of action chunks. Let
\begin{equation}
    \mathbf{a} \equiv \mathbf{a}_{t:t+H}
\end{equation}
be a ground-truth action chunk sampled from the demonstration data distribution. We sample Gaussian noise with the same shape as the action chunk,
\begin{equation}
    \boldsymbol{\epsilon} \sim \mathcal{N}(0, I),
\end{equation}
and a continuous flow time
\begin{equation}
    \tau \sim \mathcal{U}(0,1).
\end{equation}
Following the standard rectified-flow / flow-matching construction, we define a linear interpolation path between noise and data:
\begin{equation}
    \mathbf{x}_{\tau}
    =
    (1-\tau)\boldsymbol{\epsilon}
    +
    \tau \mathbf{a}.
\end{equation}
The corresponding target velocity along this path is
\begin{equation}
    \mathbf{v}
    =
    \frac{d\mathbf{x}_{\tau}}{d\tau}
    =
    \mathbf{a} - \boldsymbol{\epsilon}.
\end{equation}

The CoRE-augmented Action DiT parameterizes a conditional velocity field:
\begin{equation}
    \hat{\mathbf{v}}_{\theta}
    =
    f_{\theta,\mathcal{R}}
    \big(
        \mathbf{x}_{\tau},
        \tau,
        s_t,
        C(z_{\mathrm{mod}}),
        g,
        z_{\mathrm{mod}}
    \big),
\end{equation}
where $\theta$ denotes the model parameters and $\mathcal{R}$ denotes the learned CoRE routing mechanism. The flow-matching loss is
\begin{equation}
    \mathcal{L}_{\mathrm{FM}}
    =
    \mathbb{E}_{\mathbf{a}, I_t, \ell_{\mathrm{lang}}, s_t, \boldsymbol{\epsilon}, \tau, z_{\mathrm{mod}}}
    \left[
        \left\|
        f_{\theta,\mathcal{R}}
        \big(
            \mathbf{x}_{\tau},
            \tau,
            s_t,
            C(z_{\mathrm{mod}}),
            g,
            z_{\mathrm{mod}}
        \big)
        -
        (\mathbf{a}-\boldsymbol{\epsilon})
        \right\|_2^2
    \right].
\end{equation}

For a fixed conditioning context
\begin{equation}
    y = (I_t, \ell_{\mathrm{lang}}, s_t, z_{\mathrm{mod}}),
\end{equation}
the pointwise minimizer of the above objective is
\begin{equation}
    f^*(\mathbf{x},\tau,y)
    =
    \mathbb{E}
    \left[
        \mathbf{a}-\boldsymbol{\epsilon}
        \mid
        \mathbf{x}_{\tau}=\mathbf{x}, y
    \right].
\end{equation}
Thus, under sufficient model capacity and optimization, the learned vector field estimates the conditional transport velocity from Gaussian noise to the action-chunk distribution conditioned on visual observations, language, proprioception, and modality availability.

\subsection{CoRE-Parameterized Action Diffusion Transformer}

The noisy action chunk $\mathbf{x}_{\tau}$, flow time $\tau$, and state $s_t$ are embedded into action-side tokens. Let
\begin{equation}
    H^0
    =
    [E_s(s_t); Q_{\mathrm{act}}; E_a(\mathbf{x}_{\tau},\tau)]
    =
    \{h_i^0\}_{i=1}^{L_a}
\end{equation}
denote the initial action-side token sequence, where $E_s$ is the state encoder, $E_a$ is the noisy-action and time encoder, $Q_{\mathrm{act}}$ are learnable action query tokens, and $L_a$ is the number of action-side tokens.

A standard dense Action DiT applies the same computation path to all action-side tokens. In contrast, CoRE-VLA replaces selected transformer layers with CoRE blocks. Let $\mathcal{I}_{\mathrm{CoRE}}$ denote the set of layers equipped with CoRE. In a CoRE layer $\ell\in\mathcal{I}_{\mathrm{CoRE}}$, the input sequence is
\begin{equation}
    H^{\ell}
    =
    \{h_i^{\ell}\}_{i=1}^{L_a}.
\end{equation}
CoRE performs two levels of conditional computation: task-intent-conditioned token selection and modality-aware expert routing.

\subsection{Intent-Conditioned Action-Side Token Selection}

For each action-side token $h_i^{\ell}$, CoRE predicts a continuous selection score conditioned on the task intent:
\begin{equation}
    p_i^{\ell}
    =
    \sigma
    \left(
        w_{\mathrm{sel}}^{\top}
        [
            \operatorname{LN}(h_i^{\ell}); g
        ]
    \right),
\end{equation}
where $w_{\mathrm{sel}}$ is a learnable vector, $\operatorname{LN}(\cdot)$ denotes layer normalization, and $\sigma(\cdot)$ is the sigmoid function. Given a target capacity ratio $\rho$, CoRE selects a capacity-limited subset
\begin{equation}
    \mathcal{S}^{\ell}
    =
    \operatorname{TopK}
    \left(
        \{p_i^{\ell}\}_{i=1}^{L_a},
        \max(1,\lfloor \rho L_a \rfloor)
    \right).
\end{equation}
Only tokens in $\mathcal{S}^{\ell}$ undergo cross-attention and expert computation. Tokens outside $\mathcal{S}^{\ell}$ bypass the conditional computation in this layer:
\begin{equation}
    h_i^{\ell+1}=h_i^{\ell},
    \quad
    i\notin\mathcal{S}^{\ell}.
\end{equation}
This mechanism allows CoRE-VLA to allocate action-generation computation to the most task-relevant action-side representations while keeping the computation bounded by the capacity ratio $\rho$.

For a selected token $i\in\mathcal{S}^{\ell}$, the token first attends to the condition sequence:
\begin{equation}
    \tilde{h}_i^{\ell}
    =
    h_i^{\ell}
    +
    p_i^{\ell}
    \cdot
    \operatorname{CrossAttn}_{\ell}
    \left(
        \operatorname{LN}(h_i^{\ell}),
        C(z_{\mathrm{mod}})
    \right).
\end{equation}
The selection score $p_i^{\ell}$ acts as a continuous gate on the selected computation path. Although the Top-K decision is discrete, gradients are propagated through the continuous gates and routed expert outputs for the selected tokens.

\subsection{Modality-Aware Expert Routing}

Each CoRE layer contains a set of general experts and modality-specialized experts:
\begin{equation}
    \mathcal{E}
    =
    \mathcal{E}_{\mathrm{gen}}
    \cup
    \mathcal{E}_{\mathrm{mod}}.
\end{equation}
General experts are intended to model reusable action-processing patterns shared across tasks and sensor configurations, while modality-specialized experts model computation that depends on the auxiliary modality.

The availability mask for expert $e$ is defined as
\begin{equation}
    M_e(z_{\mathrm{mod}})
    =
    \begin{cases}
    1, & e\in \mathcal{E}_{\mathrm{gen}}, \\
    1, & e\in \mathcal{E}_{\mathrm{mod}} \ \mathrm{and}\ z_{\mathrm{mod}}=1, \\
    0, & e\in \mathcal{E}_{\mathrm{mod}} \ \mathrm{and}\ z_{\mathrm{mod}}=0.
    \end{cases}
\end{equation}
Thus, when the auxiliary modality is unavailable, modality-specialized experts are physically disabled and cannot be selected by the router.

For a selected token $i\in\mathcal{S}^{\ell}$, the expert-routing logits are
\begin{equation}
    r_{i,e}^{\ell}
    =
    (w_e^{\ell})^{\top}
    [
        \operatorname{LN}(\tilde{h}_i^{\ell}); g
    ],
    \quad
    e\in\mathcal{E},
\end{equation}
where $w_e^{\ell}$ is a learnable routing vector for expert $e$. We apply the modality-availability mask before expert selection:
\begin{equation}
    \bar{r}_{i,e}^{\ell}
    =
    \begin{cases}
    r_{i,e}^{\ell}, & M_e(z_{\mathrm{mod}})=1, \\
    -\infty, & M_e(z_{\mathrm{mod}})=0.
    \end{cases}
\end{equation}
The routing probabilities over available experts are
\begin{equation}
    \pi_{i,e}^{\ell}
    =
    \operatorname{softmax}
    (\bar{r}_{i,:}^{\ell})_e.
\end{equation}
For token-choice routing with one selected expert, the routed expert is
\begin{equation}
    e_i^{\ell}
    =
    \arg\max_{e\in\mathcal{E}}
    \bar{r}_{i,e}^{\ell}.
\end{equation}
The expert update is then
\begin{equation}
    \Delta h_i^{\ell}
    =
    p_i^{\ell}
    \cdot
    \pi_{i,e_i^{\ell}}^{\ell}
    \cdot
    E_{e_i^{\ell}}^{\ell}
    \left(
        \operatorname{LN}(\tilde{h}_i^{\ell})
    \right),
\end{equation}
and the selected token is updated by a residual connection:
\begin{equation}
    h_i^{\ell+1}
    =
    \tilde{h}_i^{\ell}
    +
    \Delta h_i^{\ell},
    \quad
    i\in\mathcal{S}^{\ell}.
\end{equation}
After all selected tokens are updated, CoRE scatters them back to their original positions and merges them with the bypassed tokens, producing the full sequence $H^{\ell+1}$.

After the final Action DiT layer, an action decoder maps the final action-side sequence to the predicted velocity:
\begin{equation}
    \hat{\mathbf{v}}_{\theta}
    =
    D_{\theta}(H^L).
\end{equation}
Therefore, CoRE-VLA can be viewed as a flow-matching action generator whose velocity field is parameterized by task- and modality-conditioned sparse computation.

\subsection{Routing Regularization}

The flow-matching objective trains the model to predict the correct action velocity. We additionally use routing regularization to stabilize sparse conditional computation.

First, the selection regularizer encourages the average selection mass to match the target capacity ratio $\rho$:
\begin{equation}
    \mathcal{L}_{\mathrm{sel}}
    =
    \sum_{\ell\in\mathcal{I}_{\mathrm{CoRE}}}
    \left(
        \frac{1}{L_a}
        \sum_{i=1}^{L_a}
        p_i^{\ell}
        -
        \rho
    \right)^2.
\end{equation}
This prevents the selector from degenerating into selecting either too many or too few action-side tokens.

Second, we use a token-choice load-balancing loss over the available experts. Let $\mathcal{B}^{\ell}$ denote the set of selected tokens in layer $\ell$ across a mini-batch, and let
\begin{equation}
    \mathcal{E}_{\mathrm{avail}}(z_{\mathrm{mod}})
    =
    \{e\in\mathcal{E}\mid M_e(z_{\mathrm{mod}})=1\}
\end{equation}
be the set of available experts. For each expert $e$, define the average routing probability
\begin{equation}
    P_e^{\ell}
    =
    \frac{1}{|\mathcal{B}^{\ell}|}
    \sum_{i\in\mathcal{B}^{\ell}}
    \pi_{i,e}^{\ell},
\end{equation}
and the empirical assignment fraction
\begin{equation}
    F_e^{\ell}
    =
    \frac{1}{|\mathcal{B}^{\ell}|}
    \sum_{i\in\mathcal{B}^{\ell}}
    \mathbf{1}[e_i^\ell = e].
\end{equation}
The load-balancing loss is
\begin{equation}
    \mathcal{L}_{\mathrm{moe}}^{\ell}
    =
    |\mathcal{E}_{\mathrm{avail}}|
    \sum_{e\in\mathcal{E}_{\mathrm{avail}}}
    P_e^{\ell}F_e^{\ell},
\end{equation}
and the total MoE regularization is
\begin{equation}
    \mathcal{L}_{\mathrm{moe}}
    =
    \sum_{\ell\in\mathcal{I}_{\mathrm{CoRE}}}
    \mathcal{L}_{\mathrm{moe}}^{\ell}.
\end{equation}

The final training objective is
\begin{equation}
    \mathcal{L}
    =
    \mathcal{L}_{\mathrm{FM}}
    +
    \lambda_{\mathrm{sel}}\mathcal{L}_{\mathrm{sel}}
    +
    \lambda_{\mathrm{moe}}\mathcal{L}_{\mathrm{moe}},
\end{equation}
where $\lambda_{\mathrm{sel}}$ and $\lambda_{\mathrm{moe}}$ are weighting coefficients.

\subsection{Adaptive Inference by Flow Integration}

At inference time, CoRE-VLA constructs the condition sequence and expert mask according to the available sensors. If the auxiliary modality is available, the model uses
\begin{equation}
    C = [Z_{\mathrm{mod}}; Z_{\mathrm{vl}}],
    \quad
    M_e=1,\ \forall e\in\mathcal{E}.
\end{equation}
If the auxiliary modality is unavailable, the model uses
\begin{equation}
    C = Z_{\mathrm{vl}},
    \quad
    M_e =
    \begin{cases}
    1, & e\in\mathcal{E}_{\mathrm{gen}}, \\
    0, & e\in\mathcal{E}_{\mathrm{mod}}.
    \end{cases}
\end{equation}
The action chunk is generated by integrating the learned velocity field from Gaussian noise to the action space. We sample
\begin{equation}
    \mathbf{x}_0 \sim \mathcal{N}(0,I),
\end{equation}
discretize $[0,1]$ into $K$ steps with $\Delta\tau=1/K$, and apply Euler integration:
\begin{equation}
    \mathbf{x}_{\tau_{k+1}}
    =
    \mathbf{x}_{\tau_k}
    +
    \Delta\tau
    f_{\theta,\mathcal{R}}
    \big(
        \mathbf{x}_{\tau_k},
        \tau_k,
        s_t,
        C(z_{\mathrm{mod}}),
        g,
        z_{\mathrm{mod}}
    \big),
    \quad
    k=0,\ldots,K-1.
\end{equation}
The final state is decoded as the predicted action chunk:
\begin{equation}
    \hat{\mathbf{a}}_{t:t+H}
    =
    \mathbf{x}_1.
\end{equation}
The predicted chunk is then executed in a receding-horizon closed-loop manner.

\subsection{Interpretation}

This formulation highlights the role of CoRE inside flow-matching VLA policies. The flow-matching objective defines what the action generator should predict: a conditional velocity field from noise to demonstrated action chunks. CoRE defines how this velocity field is parameterized: task intent selects the action-side representations that require conditional computation, while modality availability determines which experts are eligible for routing. As a result, the same policy can exploit auxiliary perceptual modalities when they are present, fall back to an RGB-language computation path when they are absent, and allocate action-generation capacity adaptively across tasks and long-horizon subgoals.

\end{document}